\documentclass[10pt, a4paper]{article}

\usepackage{multibib}
\usepackage{graphicx}
\usepackage{tabularx}
\usepackage{soul}
\usepackage{multirow}

\usepackage{epstopdf}
\usepackage[utf8]{inputenc}

\usepackage{hyperref}
\usepackage{xstring}

\usepackage{xeCJK}
\usepackage{makecell}
\usepackage{array}
\usepackage{color}

\usepackage{float}
\usepackage{booktabs}

\usepackage[most]{tcolorbox}
\tcbset{enhanced, breakable}
\tcbuselibrary{theorems}
\usepackage{hyperref}
\usepackage{lipsum}

\usepackage{placeins}

\newtcbtheorem{contract}{Contract}{
  breakable,
  enhanced,
  colback=gray!5!white,
  colframe=gray!75!black,
  fonttitle=\bfseries
}{box}

\newtcbtheorem{prompt}{Prompt}{ 
    breakable, 
    enhanced, 
    colback=gray!5!white, 
    colframe=brown!75!black, 
    fonttitle=\bfseries 
}{box}

\newtcbtheorem{summary}{Summary}{
  breakable,
  enhanced,
  colback=gray!5!white,
  colframe=orange!75!black,
  fonttitle=\bfseries
}{box}

\def\reqrev{\textrm{ReqRev}}
\def\riskrev{\textrm{RiskRev}}
\def\statrev{\textrm{StatRev}}
\def\amendexp{\textrm{AmendExp}}

\newcolumntype{Y}{>{\centering\arraybackslash}X}

\newif\ifshowjapanese
\newcommand{\jp}[1]{\ifshowjapanese{\color{blue}#1}\fi}
\showjapanesefalse

\usepackage[final]{lrec2026} 

\title{LegalRikai: Open Benchmark - A Benchmark for Complex Japanese Corporate Legal Tasks}

\name{Shogo Fujita$^1$, Yuji Naraki$^1$, Yiqing Zhu$^1$, Shinsuke Mori$^2$} 

\address{$^1$LegalOn Technologies, Inc.,~$^2$Kyoto University \\
         shogo.fujita@legalontech.jp}

\abstract{
This paper introduces LegalRikai: Open Benchmark, a new benchmark comprising four complex tasks that emulate Japanese corporate legal practices. The benchmark was created by legal professionals under the supervision of an attorney. This benchmark has 100 samples that require long-form, structured outputs, and we evaluated them against multiple practical criteria. We conducted both human and automated evaluations using leading LLMs, including GPT-5, Gemini 2.5 Pro, and Claude Opus 4.1. Our human evaluation revealed that abstract instructions prompted unnecessary modifications, highlighting model weaknesses in document-level editing that were missed by conventional short-text tasks. 
Furthermore, our analysis reveals that automated evaluation aligns well with human judgment on criteria with clear linguistic grounding, and assessing structural consistency remains a challenge. The result demonstrates the utility of automated evaluation as a screening tool when expert availability is limited. We propose a dataset evaluation framework to promote more practice-oriented research in the legal domain.
 \\ \newline \Keywords{Legal Applications, Legal Review} }

\begin{document}

\maketitleabstract

\section{Introduction}

The adoption of LLMs is rapidly increasing, and there are high expectations for their ability to streamline legal tasks such as document generation, summarization, and proofreading. However, actual corporate legal workflows are not single nor simple tasks; they consist of complex operations integrating multiple processes. The ability of LLMs to handle such complex tasks has only been evaluated within limited scopes.

\begin{figure}[t]
\includegraphics[width=\columnwidth]{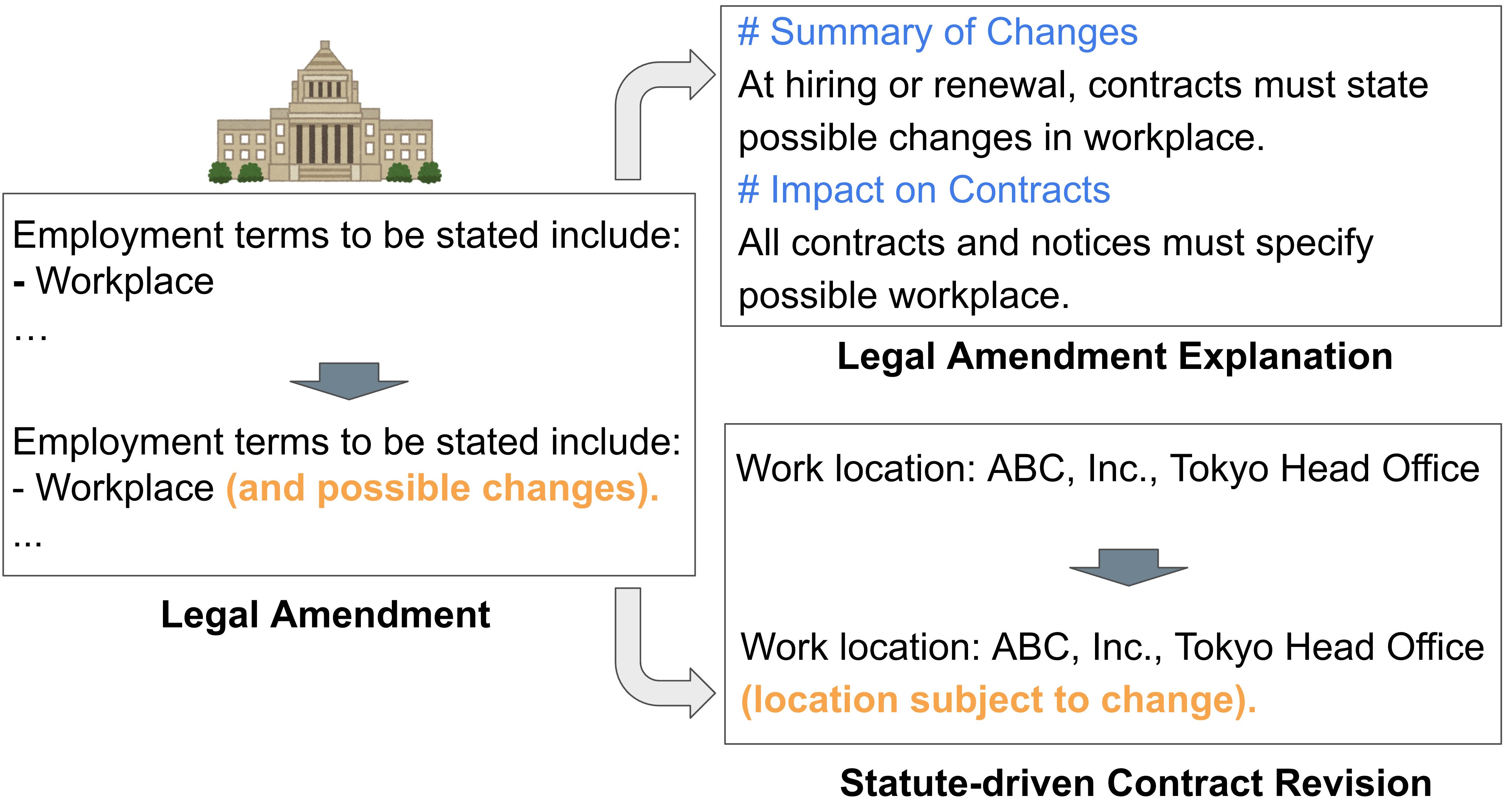}
\caption{The overview of the legal department's operations regarding a legal amendment.}
\label{intro_overview}
\end{figure}

The practical legal work targeted in this paper involves recurring editing workflows such as: (1) Internal communication regarding the business impact of legal amendments, (2) Contract revisions in response to legal amendments, (3) Requirement-driven contract revisions based on stakeholder requests, and (4) Risk-driven contract revisions aimed at preventing future disputes and disadvantages.
For example, when a law is amended, a legal department typically performs tasks (1) and (2), as illustrated in Figure \ref{intro_overview}. The former requires multiple processes: (a) understanding the pre-amendment law, (b) comprehending the differences between pre- and post-amendment, (c) summarizing the changes, and (d) listing the impact on contracts. The latter, in addition to these, requires (e) understanding the structure and content of the existing contract, (f) identifying the clauses that need editing, and (g) performing edits while preserving the original contract's format.

Existing legal benchmarks primarily focus on short-answer QA or classification tasks that test legal knowledge. They fail to adequately measure the long-form editing capabilities essential for practical work, such as modifying clauses while maintaining the overall consistency of a contract. This capability gap is critical, as contract drafts generated by LLMs may appear correct at first glance but may contain hidden flaws, such as incorrect clause references, inconsistent definitions, or contextually inappropriate outputs. Therefore, we propose a benchmark of four complex tasks modeled on real-world workflows. We employ both human and automated evaluations to analyze the outputs by leading LLMs, identifying which aspects have reached a practical level and where weaknesses are most prominent.

\paragraph{Research Questions} This study aims to answer the following questions: \begin{enumerate} 
\item \textbf{RQ1:} How do differences in legal tasks affect the editing behavior of LLMs? 
\item \textbf{RQ2:} To what extent does performance on the proposed tasks correlate with performance on conventional benchmarks?
\item \textbf{RQ3:} How closely do human and automated evaluations align in assessing editing quality, and can automated evaluation effectively substitute for human judgment?
\end{enumerate} 
The contributions of this research are as follows: 
\begin{itemize} 
\item We propose LegalRikai: Open Benchmark, a new benchmark consisting of four complex tasks directly tied to major workflows in Japanese legal practice. Legal professionals created the dataset under the supervision of qualified lawyers.
\item We conducted inference and evaluation on the benchmark using leading LLMs and analyzed their performance and tendencies. 
\item We specifically identify future research challenges regarding the adoption of LLMs in the legal domain. 
\end{itemize}

\section{Proposed Corpora}
\label{corpora_exp}

In this study, we define the four workflows described in the previous section as tasks: \textbf{Legal amendment explanation (\amendexp)} takes pre- and post-amendment statutes as input and outputs the summary of the amendment and its impact on contracts. \textbf{Statute-driven contract revision (\statrev)} takes pre- and post-amendment statutes and a contract as input, and outputs a contract that complies with the current statute. \textbf{Requirement-driven contract revision (\reqrev)} takes a contract and direct revision instructions as input, and outputs a contract edited according to those instructions. \textbf{Risk-driven contract revision (\riskrev)} takes a contract and instructions to mitigate legal risks as input, and outputs a contract edited according to those instructions. Below, we describe each task and its evaluation metrics. For each of the four tasks, we selected evaluation metrics that could be assessed both manually and automatically by LLMs. The metrics are set on a 3-point scale for cases where partial achievement needs to be distinguished, and on a 2-point scale for cases where achievement is clearly binary.

\subsection{\amendexp:~Legal amendment explanation}

Legal amendments can have wide-ranging effects on contracts and business operations. \amendexp~aims to compare pre- and post-amendment statutes, generating a summary of the changes and an explanation of their impacts on contracts. 
This task takes pre- and post-amendment statutes as input. 
This task simulates the practical work of explaining legal revisions to stakeholders and compiling materials for internal discussions.

An example of this task is shown in Table \ref{amendexp_exp}. An amendment to the Penal Code introduced a new "imprisonment" (kōkinkei), necessitating changes to terms such as "imprisonment" (chōeki) and "imprisonment without work" (kinko) in contracts. These terms need to be replaced with "imprisonment" (kōkinkei).

We evaluated this task based on two main aspects: the summary of the amendment and its impact on contracts. The evaluation criteria for the amendment summary are as follows:\newline
\textbf{Coverage of Amendments (CA):} Assesses whether the model covers the main points of the change in the summary.
\textbf{Accuracy of Amendments (AA):} Assesses whether the model describes the changes accurately in the summary.
\textbf{Relevance of Amendments (RA):} Assesses whether the model excludes unnecessary content from the summary.
For the impact on contracts, as follows:
\textbf{Coverage of Impacts (CI):} Assesses whether the model addresses all relevant aspects that impact contracts in the explanations.
\textbf{Accuracy of Impacts (AI):} Assesses whether the model describes the impacts accurately in the explanations.
\textbf{Relevance of Impacts (RI):} Assesses whether the model excludes irrelevant information from the explanations.
We rated all metrics for \amendexp~on a 3-point scale.

\begin{table*}[htbp]
\centering
\footnotesize
\begin{tabularx}{\textwidth}{lX}
\hline
Pre-amendment statute &
\jp{...2~~懲役は、刑事施設に拘置して所定の作業を行わせる。...} 
...(2)"imprisonment" (chōeki) consists of confinement in a penal institution with assigned work....\\
\hline
Post-amendment statute &
\jp{...2~~拘禁刑は、刑事施設に拘置する。3~~拘禁刑に処せられた者には、改善更生を図るため、必要な作業を行わせ、又は必要な指導を行うことができる。 ...}
...(2)"imprisonment" (kōkinkei) shall consist of confinement in a penal institution. (3)A person sentenced to "imprisonment" (kōkinkei) may be required to perform necessary labor or be provided with necessary guidance for their rehabilitation and reintegration. ...\\
\hline\hline
Ground truth (Summary of change) &
\jp{刑事施設における受刑者の処遇及び執行猶予制度等のより一層の充実を図るため、懲役及び禁錮を廃止して拘禁刑を創設。 }
The sentences of "imprisonment" (chōeki) and "imprisonment without work" (kinko) have been abolished and replaced with a newly established, unified sentence of "imprisonment" (kōkinkei). This change aims to improve the treatment of inmates in penal institutions and enhance systems such as the suspension of sentences.\\
\hline
Ground truth (Impact on contracts) &
\jp{ストックオプション割当契約等、個人に何らかの権利を付与する契約において、その個人が刑罰に処せられた場合はその権利を喪失する旨が定められている場合がある。その規定において、"懲役""禁錮"という文言が使われていたら、"拘禁刑"に改正する必要がある。}
In contracts that grant certain rights to an individual, such as stock option agreements, there may be a provision stipulating the forfeiture of those rights if the individual is convicted of a criminal offense. If such a provision uses the terms "imprisonment" (chōeki) or "imprisonment without work" (kinko), it must be amended to the new term "imprisonment" (kōkinkei)\\
\hline
\end{tabularx}
\caption{An example of \amendexp.}
\label{amendexp_exp}
\end{table*}

\subsection{\statrev:~Statute-driven contract revision}
\jp{
契約書は関係者の都合だけでなく、法改正によって変更する必要が生じることもある。
本タスクは、改正前の法令に準拠した契約書および法令改正の差分情報を入力として、契約書を現行法令に適合させることを目的とする。
具体的には、改正法令に基づき修正が必要な契約条項を特定し、改訂案を生成・適用する一連の過程を含む。
\statrevは、改正法令の正確な理解と契約書条項との対応付けを要求するため、法務知識と条文解釈力を必要とする。
このタスクは、法改正直後に契約書を更新する作業を模しており、他タスクと比べて高い専門性を要する。}

Contracts often require revision due to legal amendments. 
\statrev~aims to create a contract compliant with post-amendment statutes. 
This task takes a contract that is compliant with pre-amendment statutes, as well as pre- and post-amendment statutes, as input. 
Specifically, this task involves identifying contract clauses that require revision based on the amended statute and generating and applying the revised text. 
\statrev~requires the ability to understand the amended statute and to map it to contract clauses, demanding legal knowledge and statutory interpretation. 
This task simulates the practical work of updating of contracts after legal amendments, requiring a higher level of expertise than the other tasks.

An example of this task is shown in Table \ref{statrev_exp}. This example illustrates a case where the amendment to the Labor Standards Act Enforcement Regulations necessitated the explicit definition of the scope of changes to the workplace and the duty clause.

We evaluated this task on five criteria: 
\textbf{Instruction Following (IF):} Assesses whether the model strictly performs the revision as instructed.
\textbf{Structural Consistency (SC):} Assesses whether the model maintains the resulting contract structure consistently and logically sound after the revision.
\textbf{Change Precision (CP):} Assesses whether the model makes any unnecessary revisions, deletions, or additions beyond the scope of the required change.
\textbf{Terminology Accuracy (TA):} Assesses whether the model uses the correct technical or specialized terminology throughout the revised contract.
\textbf{Wording Appropriateness (WA):}  Assesses whether the model retains appropriate contractual phrasing in the revised contract.
We rated IF, TA, and WA on a 3-point scale and SC, CP on a 2-point scale.

\begin{table*}[htbp]
\centering
\footnotesize
\begin{tabularx}{\textwidth}{lX}
\hline
Pre-amendment statute &
\jp{...第五条~~使用者が法第十五条第一項前段の規定により労働者に対して明示しなければならない労働条件は、次に掲げるものとする。 (略)一の三~~就業の場所及び従事すべき業務に関する事項...}...Article 5 (1) The working conditions which the employer shall clearly indicate to the worker under the provisions of the first sentence of paragraph....(i)-3 Matters concerning workplace and work engaged in...\\
\hline
Post-amendment statute&
\jp{...第五条~~使用者が法第十五条第一項前段の規定により労働者に対して明示しなければならない労働条件は、次に掲げるものとする。 (略)一の三~~就業の場所及び従事すべき業務に関する事項 (就業の場所及び従事すべき業務の変更の範囲を含む。) ...}...Article 5 (1) The working conditions which the employer shall clearly indicate to the worker under the provisions of the first sentence of paragraph....(i)-3 Matters concerning workplace and work engaged in (including the scope of changes to the place of work and duties to be engaged in)...\\
\hline
Input contract&
\jp{...従業員は、下記の就業場所において使用者の指示に従い誠実に行う。 就業場所:雇入れ直後　東京本社 ...}...The employee shall faithfully perform their duties in accordance with the employer's instructions at the place of work specified below. Place of work: Immediately after hiring: Tokyo Head Office...
\\
\hline\hline
Ground truth (Revised contract)&
\jp{...従業員は、下記の就業場所において使用者の指示に従い誠実に行う。 就業場所:雇入れ直後　東京本社 (変更の範囲:会社の別途指定する就業場所)...}
...The employee shall faithfully perform their duties in accordance with the employer's instructions at the place of work specified below. Place of work: Immediately after hiring: Tokyo Head Office (Scope of change: Any place of work separately designated by the Company)...
\\
\hline
\end{tabularx}
\caption{An Example of \statrev.}
\label{statrev_exp}
\end{table*}

\subsection{\reqrev:~Requirement-driven contract revision}

Contracts often require revision not only due to legal amendments but also because of requests from the parties or changes in transaction terms. 
\reqrev~aims to generate a revised contract that accurately reflects all requests. 
This task takes a contract template and multiple revision requests as input.
The revision requests range from simple wording changes to large-scale revisions spanning multiple clauses and may include instructions that do not necessarily require legal knowledge. 
The task requires identifying the target clauses, generating revised text, and maintaining overall consistency. This task is less abstract than others, as it requires faithfully reflecting explicitly given revision requests. 
This task simulates the practical work of incorporating requests from a counterparty or making wording changes during contract negotiations.

An example of this task is shown in Table \ref{reqrev_riskrev_exp}. This example has instructions with specific and explicit wording for particular clauses.
We evaluated this task on the five criteria as \statrev: IF, SC, CP, TA, and WA. 

\subsection{\riskrev:~Risk-driven contract revision}

Contracts contain potential risks that could lead to future disputes or disadvantages. 
\riskrev~aims to generate a revised contract with minimal legal risks.
This task takes a contract template and multiple revision requests designed to mitigate legal risks as input.
Some cases include instructions that do not require revision. This task contributes to pre-contractual reviews and risk reduction during negotiations.
In all cases, the input contract for \riskrev~and \reqrev~is identical. The difference is that \reqrev~provides clear instructions on how to revise, whereas \riskrev~provides high-level instructions to revise based on a stated risk. Therefore, \riskrev~requires risk identification and the formulation of revised text based on legal knowledge. 
\riskrev~ simulates the practical work of reviewing by legal staff and supporting negotiations.

An example of this task is shown in Table \ref{reqrev_riskrev_exp}. 
While \reqrev~specifies the exact wording to be changed, \riskrev~provides a legally oriented instruction, such as "revise to be more favorable to the contractor."
We evaluated this task on the same five criteria as \statrev: IF, SC, CP, TA, and WA.

\begin{table*}[htbp]
\centering
\footnotesize
\begin{tabularx}{\textwidth}{lX}
\hline
Instruction in \reqrev&
\jp{第8条 (再委託)を"委託者の事前の書面による承諾を得た場合に限り、再委託可能とする"旨に修正してください。}Please amend Article 8 (Subcontracting) to state that "subcontracting is permissible only with the prior written consent of the Consignor."\\
\hline
Instruction in \riskrev&
\jp{再委託に関する条項について、契約書内に該当するテーマや項目が含まれている場合は、当該表現を委託側に有利に修正してください。削除することが有利になる場合は、削除しても良いです。}Regarding the clause on subcontracting, if the contract contains a corresponding theme or item, please amend the expression to be more favorable to the Consignor. If deleting the clause is more advantageous, you may do so.\\
\hline
Input Contract&
\jp{...第8条 (再委託) 1.~乙は、乙の責任において、本委託業務の一部を第三者に再委託することができる。ただし、乙は、甲が要請した場合、再委託先の名称及び住所等を甲に報告しなければならない。...}...Article 8 (Subcontracting) 1. Party B may, at its own responsibility, subcontract a portion of the consigned duties to a third party. However, Party B must report the name and address of the subcontractor to Party A upon request. ...\\
\hline\hline
Ground Truth (Revised Contract)&
\jp{...第8条 (再委託) 1.~乙は、甲の事前の書面による承諾を得た場合に限り、乙の責任において、本委託業務の一部を第三者に再委託することができる。...}...Article 8 (Subcontracting) 1. Party B may, at its own responsibility, subcontract a portion of the consigned duties to a third party only upon obtaining the prior written consent of Party A. ...\\
\hline
\end{tabularx}
\caption{Examples of \reqrev~and \riskrev.}
\label{reqrev_riskrev_exp}
\end{table*}

\begin{table*}[htbp]
\centering
\begin{tabularx}{\textwidth}{lXXXX}
\hline
 & \reqrev & \riskrev & \statrev & \amendexp \\
\hline
Number of samples & 
25 & 25& 25&25\\
\hline
Avg. char. count in input contract&
6,530&6,530&9,412&-\\
Avg. articles in input contract&
16&16&38&-\\
\hline
Avg. char. count in pre-amendment statute&
-&-&27,033&32,149\\
Avg. provisions in pre-amendment statute&
-&-&292&318\\
\hline
Ground truth char. count &
253&228&300&397\\
Char. count of statute amendment diff. &
-&-&3,122&4,924\\
\hline

\end{tabularx}
\caption{Statistics for the four tasks. The ground truth char. counts for \reqrev, \riskrev, and \statrev~are the character count of the difference between the input and ground truth contracts. For \amendexp, it is the character count of the ground-truth summary and the impact on contracts.}
\label{task_list}
\end{table*}

\subsection{Datasets}

We created 25 data instances for each task. Legal professionals created the data under the supervision of an attorney employee. We created the dataset used in \riskrev, \reqrev, and \statrev~based on contracts and templates for actual practice. The statutes in \statrev~and \amendexp~are based on actual Japanese statutes.
We show statistical information in Table \ref{task_list}. All tasks require handling over 6,500 characters of information. Among them, \statrev~and \amendexp~have significant average character counts for the input statutes. Since \statrev~requires referring to 27,000 characters of statute and \amendexp~23,000 characters, long-context processing capability has a significant impact.\footnote{The dataset used in this research will be available upon application and agreement to the terms of use. Due to formatting issues with the data used in this experiment, there are several differences compared to the data that will actually be published.}

\section{Experiments}

For each of the four tasks, we conducted a human evaluation experiment and an automated evaluation experiment. In the human evaluation experiment, we performed inference with three models—GPT-5\footnote{The Model version is gpt-5-2025-08-07.}\citep{OpenAI2025GPT5}, Gemini 2.5 Pro\footnote{The Model version is gemini-2.5-pro.}\citep{google2025gemini25pro}, and Claude Opus 4.1\footnote{The Model version is claude-opus-4-1-20250805.}\citep{anthropic2025claudeopus41}—and had legal professionals evaluate the outputs under the supervision of an attorney employee.
In the automated evaluation experiment, we similarly evaluated three models—GPT-5, Gemini 2.5 Pro, and Claude Opus 4.1—as the evaluation models.

We used a task-specific template that specified the output format and constraints for each task. We provided contracts in Markdown format, and statutes in text format, converted from information obtained from the API published by \citet{DigitalAgency2025LawsAPI}\footnote{For articles whose input length exceeded the maximum context length of the LLM, the portions at the end that contained no differences between the pre- and post-amendment versions were truncated to maintain input consistency and to ensure a common input range across all LLMs.}. For \amendexp, the model outputted the summary of amendments and impact on contracts in Markdown. For \statrev, the model only outputted the clauses requiring changes in Markdown due to context length limitations. For \reqrev~and \riskrev, the model outputted the entire edited contract in Markdown.

For each model, we set the temperature to 0.0. We used GPT-5 directly via the official API, while we used Gemini 2.5 Pro and Claude Opus 4.1 via the Vertex AI Platform.

\subsection{Human Evaluation Experiment}

We conducted inference for each task using GPT-5, Gemini 2.5 Pro, and Claude Opus 4.1, and legal professionals conducted a human evaluation of each output. Each sample was independently scored by two legal professionals using the metrics defined in Section \ref{corpora_exp} (\amendexp: CA/AA/RA, CI/AI/RI; other three tasks: IF/SC/CP/TA/WA). 
When scores differed, we used the arithmetic mean. We reported Inter-annotator agreement using Cohen's $\kappa$ coefficient in Table \ref{annotator_coh_all}. Note that for metrics where all samples received the same label, $\kappa$ score is \texttt{nan}. Overall, we observed high agreement, but $\kappa$ score for WA in \reqrev~and \riskrev~were low, indicating that judgments on appropriate contractual phrasing vary considerably among individuals.
We mapped the scores to an interval scale and then normalized them to [0, 1] (Table \ref{human-all-task-result}).

As shown in Table \ref{human-all-task-result}, in \amendexp, Gemini 2.5 Pro performed best overall, with scores of AA (0.86), AI (0.54), RI (0.44), and RA (0.80). GPT-5 exhibits high coverage with CA (0.66) and CI (0.52), but its relevance is weaker, with RA (0.76) and RI (0.22), falling short of Gemini 2.5 Pro. Claude Opus 4.1 did not stand out in any metric compared to the other models.

In \statrev, Gemini 2.5 Pro scored highest in IF (0.73) and CP (0.44), excelling in coverage and precision. Claude Opus 4.1 scored highest in SC (0.40) and WA (1.00), demonstrating a strength in maintaining the proper format of a contract. However, it also had the lowest CP score (0.20), indicating that it often made extra changes not specified in the instructions. GPT-5's performance was intermediate between the other two. All models scored TA (1.00), showing no issues with the use of technical terminology.

In \reqrev, all models scored similarly on IF (0.85), suggesting that when "what to change" is explicitly stated, there is no significant difference in the ability to follow instructions. There were no significant differences in SC, TA, or WA, but GPT-5 scored CP (0.58), indicating an exceptionally high number of unnecessary revisions compared to the other models.

In \riskrev, the trends were generally similar to those in \reqrev. Since \reqrev~and \riskrev~have the same input/output format, they are likely to produce similar trends. While there was little difference in IF and SC, GPT-5 tended to perform lower than the other two models in SC, TA, and WA. GPT-5's CP (0.22) score was especially low, indicating a tendency for unnecessary revisions.

A general trend emerging from the experimental results is that the specificity of instructions has a significant influence on model behavior. In \reqrev, where we provided specific revision points and wording, all models achieved high scores in IF, demonstrating their ability to faithfully execute instructions. In contrast, in \statrev, which requires interpreting amended statutes, and \riskrev, which demands abstract risk mitigation, model performance varied greatly, with a notable increase in unnecessary modifications. The results suggest that when models attempt to interpret and supplement abstract parts of instructions, they tend to perform excessive edits or make changes that deviate from the practitioner's intended meaning. This behavior, while seemingly producing beneficial modifications, can actually increase review costs and introduce unintended legal risks.

Furthermore, model-specific characteristics became apparent. Gemini 2.5 Pro shows strength in accuracy, while Claude Opus 4.1 excelled at maintaining contract format. The results could reflect the fundamental characteristics of each model: the former specializes in careful, fact-based text generation, while the latter is adept at preserving the format and stylistic consistency of the entire document. GPT-5 demonstrated intermediate performance, exhibiting high versatility but lacking outstanding capabilities in any specific aspect. These results indicate that when selecting an LLM for legal practice, it is necessary to consider not only a single score but also the nature of the task, such as the specificity of instructions, and the model's particular strengths.

\begin{table}[t] %
\centering
\small
\begin{tabularx}{\columnwidth}{YYYYYY}
\hline
CA & AA & RA & CI & AI & RI \\
\hline
1.00 & 1.00 & 1.00 & 0.98 & 0.98 & 0.91 \\
\hline
\end{tabularx}
\vspace{1ex} 
\begin{tabularx}{\columnwidth}{lYYYYY}
\hline
Task & IF & SC & CP & TA & WA \\
\hline
\reqrev  & 0.78 & 1.00 & 0.74 & 1.00 & 0.49 \\
\riskrev & 0.78 & 1.00 & 0.69 & 0.53 & 0.60 \\
\statrev & 1.00 & 0.90 & 1.00 & nan  & 1.00 \\
\hline
\end{tabularx}

\caption{Cohen's $\kappa$ coefficient for the evaluation results of two evaluators per task. Top row: \amendexp; Bottom row: \reqrev, \riskrev, \statrev. For the TA of \statrev, the evaluation results were identical in all cases, hence marked as nan.}
\label{annotator_coh_all}
\end{table}

\begin{table}[t]
\centering
\small
\begin{tabularx}{\columnwidth}{lYYYYYY}
\hline
Model & CA & AA & RA & CI & AI & RI \\
\hline
GPT-5   & 0.66 & 0.74 & 0.76 & 0.52 & 0.44 & 0.22 \\
Gemini  & 0.64 & 0.86 & 0.80 & 0.52 & 0.54 & 0.44 \\
Claude  & 0.64 & 0.68 & 0.70 & 0.49 & 0.49 & 0.32 \\
\hline
\end{tabularx}
\vspace{1ex}
\begin{tabularx}{\columnwidth}{llYYYYY}
\hline
Task & Model & IF & SC & CP & TA & WA \\
\hline
\multirow{3}{*}{\statrev} & GPT-5   & 0.69 & 0.36 & 0.32 & 1.00 & 0.94 \\
                         & Gemini  & 0.73 & 0.36 & 0.44 & 1.00 & 0.96 \\
                         & Claude  & 0.57 & 0.40 & 0.20 & 1.00 & 1.00 \\
\hline
\multirow{3}{*}{\reqrev}  & GPT-5   & 0.85 & 0.92 & 0.58 & 0.94 & 0.79 \\
                         & Gemini  & 0.85 & 0.80 & 1.00 & 1.00 & 0.90 \\
                         & Claude  & 0.85 & 0.92 & 0.96 & 1.00 & 0.87 \\
\hline
\multirow{3}{*}{\riskrev} & GPT-5   & 0.69 & 0.88 & 0.22 & 0.79 & 0.59 \\
                         & Gemini  & 0.61 & 0.96 & 0.40 & 0.95 & 0.78 \\
                         & Claude  & 0.65 & 1.00 & 0.56 & 0.96 & 0.75 \\
\hline
\end{tabularx}

\caption{Manual evaluation results for LLM outputs. Top row: \amendexp; bottom row: \statrev, \reqrev, \riskrev. All have their value ranges changed to [0,1].}
\label{human-all-task-result}
\end{table}

\subsection{Automated Evaluation Experiment}

In this section, we conducted an automated evaluation to verify whether they can replace or assist legal professionals in their annotation work. We measured Spearman's correlation coefficient in Table \ref{all_spearman} and Mean Absolute Error (MAE) in Table \ref{all_mae} between the automated scores and human evaluations for the criteria defined in Section \ref{corpora_exp}.

We used GPT-5, Gemini 2.5 Pro, and Claude Opus 4.1 as evaluator models, setting the temperature to 0 for deterministic scoring. We provided the task definition, input, model output, and scoring criteria for each aspect, and required a structured output consisting only of the criterion name and the rationale. We also experimented with a score, Avg, which is the arithmetic mean of the scores from the three models.

As shown in Table \ref{all_spearman}, in \amendexp, a significant positive correlation was observed overall. 
The results indicate that judgments with easily demonstrable linguistic evidence, such as identifying differences in amendments and organizing their practical impacts, are well-suited for automated scoring by LLMs.
As shown in Table \ref{all_mae}, there is a difference between the summary of amendments (CA, AA, RA) and the impact on contracts (CI, AI, RI). The metrics associated with the more complex impact on contracts generally show a larger deviation from human evaluation.

In \statrev, as shown in Table \ref{all_spearman}, there are moderate correlations for IF and CP, while weak correlations for SC and WA. 
We were unable to obtain a correlation coefficient for TA because all human evaluations were identical.
The MAE scores for TA ranged from 0.28 to 0.79 in Table \ref{all_mae}, which is comparable to IF and CP, suggesting a certain level of agreement with human evaluation.

In \reqrev, as shown in Table \ref{all_spearman}, there are moderate correlations for IF and WA, while weak correlations for SC and CP, and a weak negative correlation for TA. The results are likely due to the fact that simple instructions, such as explicit replacement or addition, tend to cluster densely in human scores for each model, making it challenging to establish a rank correlation.

In \riskrev, as shown in Table \ref{all_spearman}, there are moderate correlations for CP and WA, and weak correlations for IF, SC, and TA. Since this task has a high level of instruction abstraction, evaluating IF was likely the most difficult among the four tasks.

A general trend is that for tasks like \amendexp, where contained linguistic evidence in the text, there was a significant correlation between automated and human evaluations. However, the markedly weak correlation in contract revision tasks, especially for SC and TA, highlights the limitations of LLM-based evaluation. 

The difficulty in evaluating SC is likely due to LLMs processing text as a linear sequence of tokens. Human evaluators perceive the visual and logical structure of a document as a whole, such as clause number references and indentation breaks. In contrast, LLM evaluators are adept at assessing local semantic meaning but struggle to accurately track the consistency of global references, such as "Article 15 refers to Article 9."

The discrepancy in TA evaluation is rooted in the nuanced context of legal terminology. For example, while "immediately" and "promptly" are similar, their levels of obligation differ in legal documents. Human experts judge these nuances, whereas LLM evaluators tend to assess based on dictionary correctness or word frequency, failing to evaluate the appropriateness of term choice in context. From these considerations, it is evident that automated evaluation is practical for screening aspects related to content, such as coverage and instruction following. However, for assessing formal and contextual validity, such as document structure and professional nuance, supervision by human experts remains necessary. Furthermore, the Avg score tended to improve correlation by offsetting systemic biases, proving effective in achieving a more accurate evaluation.

\begin{table}[t]
\small
\centering
\begin{tabularx}{\columnwidth}{lYYYYYY}
\hline
   Model    & CA    & AA    & RA    & CI    & AI    & RI    \\
\hline
 GPT-5   & 0.72$^\dagger$ & 0.32$^\dagger$  & 0.41$^\dagger$  & 0.50$^\dagger$ & 0.44$^\dagger$ & 0.23$^\dagger$  \\
 Gemini & 0.58$^\dagger$ & 0.36$^\dagger$ & 0.47$^\dagger$ & 0.36$^\dagger$ & 0.31$^\dagger$ &  0.13 \\
 Claude & 0.53$^\dagger$ & 0.37$^\dagger$ & 0.46$^\dagger$ & 0.53$^\dagger$ & 0.54$^\dagger$ &  0.41$^\dagger$ \\
 Avg & 0.68$^\dagger$ & 0.38$^\dagger$ & 0.51$^\dagger$ & 0.55$^\dagger$ & 0.47$^\dagger$ & 0.31$^\dagger$ \\
\hline
\end{tabularx}
\vspace{1ex} 
\begin{tabularx}{\columnwidth}{llYYYYY}
\hline
Task & Model & IF & SC & CP & TA & WA \\
\hline
\multirow{4}{*}{\statrev} & GPT-5   & 0.34$^\dagger$ & 0.07 & 0.47$^\dagger$ & nan & \!\!-0.04\!\! \\
                         & Gemini  & 0.43$^\dagger$ & 0.11 & 0.46$^\dagger$ & nan & 0.08 \\
                         & Claude  & 0.34$^\dagger$ & 0.08 & 0.58$^\dagger$ & nan & 0.19 \\
                         & Avg     & 0.49$^\dagger$ & 0.12 & 0.65$^\dagger$ & nan & 0.14 \\
\hline
\multirow{4}{*}{\reqrev}  & GPT-5   & 0.17 & 0.09 & 0.15 & \!\!-0.05\!\! & 0.00 \\
                         & Gemini  & 0.10 & 0.08  & 0.17 & \!\!-0.06\!\! & 0.27$^\dagger$ \\
                         & Claude  & 0.26$^\dagger$ & 0.16 & nan & \!\!-0.12\!\! & 0.07 \\
                         & Avg     & 0.23$^\dagger$ & 0.12 & 0.18 & \!\!-0.15\!\! & 0.23$^\dagger$ \\
\hline
\multirow{4}{*}{\riskrev} & GPT-5   & 0.16 & \!\!-0.08\!\! & 0.36$^\dagger$ & 0.14 & 0.22 \\
                         & Gemini  & \!\!-0.13\!\! & 0.01 & 0.26$^\dagger$ & \!\!-0.02\!\! & 0.20 \\
                         & Claude  & 0.20 & \!\!0.07\!\! & 0.35 & 0.11 & 0.26$^\dagger$ \\
                         & Avg     & 0.11 & \!\!-0.07\!\! & 0.41$^\dagger$ & 0.19 & 0.34$^\dagger$ \\
\hline
\end{tabularx}

\caption{Spearman correlation coefficients between automated and manual evaluations. Top row: \amendexp; Bottom row: \statrev, \reqrev, \riskrev.
$^\dagger$ denotes a statistically significant difference from manual evaluation ($p<0.05$).}
\label{all_spearman}
\end{table}

\begin{table}[t]
\small
\centering
\begin{tabularx}{\columnwidth}{lYYYYYY}
\hline
  Model     & CA    & AA    & RA    & CI    & AI    & RI    \\
\hline
GPT-5   & 0.31 & 0.47 & 0.71 & 0.66 & 0.66 & 0.64 \\
Gemini  & 0.39 & 0.44 & 0.47 & 0.69 & 0.75 & 1.01 \\
Claude  & 0.44 & 0.53 & 0.56 & 0.58 & 0.55 & 0.63 \\
Avg     & 0.37 & 0.47 & 0.58 & 0.58 & 0.60 & 0.71 \\
\hline
\end{tabularx}
\vspace{1ex} 
\begin{tabularx}{\columnwidth}{llYYYYY}
\hline
Task & Model & IF & SC & CP & TA & WA \\
\hline
\multirow{4}{*}{\statrev} & GPT-5   & 0.54 & 0.60 & 0.24 & 0.52 & 0.31 \\
                         & Gemini  & 0.58 & 0.57 & 0.25 & 0.28 & 0.36 \\
                         & Claude  & 0.75 & 0.57 & 0.23 & 0.79 & 0.59 \\
                         & Avg     & 0.56 & 0.58 & 0.24 & 0.53 & 0.41 \\
\hline
\multirow{4}{*}{\reqrev}  & GPT-5   & 0.33 & 0.15 & 0.21 & 0.09 & 0.31 \\
                         & Gemini  & 0.35 & 0.21 & 0.19 & 0.13 & 0.28 \\
                         & Claude  & 0.42 & 0.27 & 0.17 & 0.17 & 0.28 \\
                         & Avg     & 0.34 & 0.21 & 0.17 & 0.17 & 0.28 \\
\hline
\multirow{4}{*}{\riskrev} & GPT-5   & 0.43 & 0.15 & 0.33 & 0.25 & 0.52 \\
                         & Gemini  & 0.62 & 0.27 & 0.42 & 0.24 & 0.52 \\
                         & Claude  & 0.46 & 0.43 & 0.38 & 0.24 & 0.48 \\
                         & Avg     & 0.49 & 0.28 & 0.36 & 0.23 & 0.49 \\
\hline
\end{tabularx}

\caption{Mean Absolute Error(MAE) between manual and automated evaluations. Top row: \amendexp ; Bottom row: \statrev, \reqrev, \riskrev}
\label{all_mae}
\end{table}

\subsection{Performance Variation by LLM Scale}

We investigated how model scale affects performance on document-level editing tasks. Table \ref{amend-exp-GPT-5s} compares the results of the full-GPT-5 model with its lightweight versions, GPT-5-mini\footnote{The Model version is gpt-5-mini-2025-08-07} and GPT-5-nano\footnote{The Model version is gpt-5-nano-2025-08-07.}.
As shown in Table \ref{amend-exp-GPT-5s}, reducing the model scale resulted in a significant performance drop, especially in aspects that require understanding the entire document's context, such as CA and CI. For instance, the CI score dropped by 0.25 from 0.52 for the full model to 0.27 for the nano model.
To verify whether this rate of performance degradation is specific to our proposed tasks, we compare it with the benchmark results published \citep{OpenAI2025GPT5}. 
In the GPQA diamond benchmark, which measures graduate-level question-answering ability, the performance drop was only 14.5\%, with the score decreasing from 85.7\% for GPT-5 to 71.2\% for GPT-5-nano.
This comparison reveals that the reduction in model scale leads to significantly more severe performance degradation in specialized, long-text reading and editing tasks, such as those in our study, compared to the performance difference observed in general-purpose reasoning tasks. The results suggest that the ability to perform tasks while retaining complex information, such as the overall structure of a contract and inter-clause references, is strongly dependent on model scale.
Therefore, although lightweight models may be attractive in terms of cost and speed, applying them carelessly to tasks where document-wide consistency is critical, such as contract review, carries a high risk of inducing fatal oversights. 
The results emphasize the importance of selecting a model of appropriate scale based on the task's difficulty.

\begin{table}[t]
\centering
\small
\begin{tabularx}{\columnwidth}{lYYYYYY}
\hline
Model &CA&AA&RA&CI&AI&RI\\
\hline
GPT-5&0.66&0.74&0.76&0.52&0.44&0.22\\
GPT-5-mini&0.63&0.83&0.55&0.37&0.42&0.37\\
GPT-5-nano&0.51&0.65&0.44&0.27&0.28&0.27\\
\hline
\end{tabularx}
\caption{Evaluation results for \amendexp~using each GPT-5 model by Avg evaluator. All have their value ranges changed to
[0,1].}
\label{amend-exp-GPT-5s}
\end{table}

\section{Related work}

With the advancement of LLMs, research on natural language processing tasks in the legal domain has become active. In particular, numerous benchmark datasets have been proposed to evaluate the legal knowledge and reasoning abilities of LLMs, with diverse designs corresponding to the legal systems and document formats of various countries and languages. 
Below, we organize prior research from three perspectives relevant to our study.

Several benchmarks have been proposed to evaluate the legal knowledge and reasoning ability of LLMs. Comprehensive surveys and datasets such as MMLU \citeplanguageresource{MMLU2021}, LawBench \citeplanguageresource{fei-etal-2024-lawbench}, LegalBench \citeplanguageresource{guha2023legalbench}, LegalAgentBench \citeplanguageresource{li2024legal_agent_bench}, and LexGLUE \citeplanguageresource{chalkidis-etal-2022-lexglue} cover diverse aspects of legal understanding, from factual recall to multi-hop reasoning and case law analysis \citep{zhong-etal-2020-nlp}.
LEXam \citeplanguageresource{fan2025lexam} is built from 340 law exams across 116 law school courses, spanning multiple subjects and degree levels.
BriefMe \citeplanguageresource{woo-etal-2025-briefme} focuses on trial preparation tasks such as argument summarization, completion, and case retrieval.
In addition, multi-task benchmarks based on non-English legal systems, including those of South Korea and China, have also been introduced \citeplanguageresource{NEURIPS2022_korean,LeCaRD,LeCaRDv2}.

On the other hand, benchmarks specific to the Japanese legal domain have not received sufficient attention. Therefore, in the next section, we will review the current state of LLM evaluation in Japanese, as well as datasets for the legal domain. 

\subsection{Benchmark Datasets in Japanese}

Recent studies have developed diverse benchmarks for evaluating Japanese LLMs. The llm-jp-eval platform \citep{llmjp2024} enables unified automatic evaluation across tasks such as classification, summarization, and reasoning, while judge-based and judge-free approaches, including Japanese Vicuna QA, Japanese MT-Bench, and Judge-free Benchmark \citeplanguageresource{sun-etal-2024-rapidly,Imos2025-judge-free}, leverage LLMs themselves for scalable assessment.
Domain-specific benchmarks have also emerged, such as FinBench for finance \citeplanguageresource{Hirano2023-finnlpkdf}, JMedBench for medicine \citeplanguageresource{jiang-etal-2025-jmedbench}, JFLD for formal logic reasoning \citeplanguageresource{morishita-etal-2024-jfld}, and LCTG Bench for controllable generation \citeplanguageresource{kurihara2025lctgbenchllmcontrolled}. Moreover, LawQA-JP \citeplanguageresource{digitalagency_lawqajp} tests understanding of Japanese laws and regulations.
Despite such progress, no benchmark yet captures the practical context of Japanese legal work. To address this gap, we construct a benchmark reflecting real-world tasks of Japanese legal professionals.

\subsection{Complex Task Benchmarks and Evaluation Design} 

In LLM evaluation, complex tasks involving multiple reasoning steps or processes, rather than single knowledge recall or classification, are gaining attention. The trend is relevant to the design of tasks like contract review, which require multi-stage processing such as "understanding document structure," "interpreting clause meaning," and "risk identification or legal judgment." 
Recent benchmarks have explored LLMs’ practical and reasoning capabilities through diverse complex tasks. GAIA \citeplanguageresource{GAIA_2023} and Humanity’s Last Exam \citeplanguageresource{phan2025humanitysexam} evaluate general and expert-level reasoning, while SWE-bench \citeplanguageresource{jimenez2024swebench} and PaperBench \citeplanguageresource{starace2025paperbench} assess multi-stage problem solving in realistic settings such as software debugging and paper reproduction.
Benchmarks like ProcBench, ToolComp, and Multi-LogiEval \citeplanguageresource{fujisawa2024procbench,nath2025toolcomp,patel-etal-2024-multi} further emphasize the importance of stepwise reasoning and tool use.

These studies have laid the groundwork for a multifaceted understanding of LLM capabilities, ranging from knowledge recall to the evaluation of complex reasoning processes. Existing legal benchmarks often remain confined to assessing knowledge of statutes and case law, or limited reasoning tasks, lacking tasks that require complex reasoning processes. We have addressed this challenge.

\section{Conclusion}

This paper proposes LegalRikai: Open Benchmark, a new benchmark consisting of four complex tasks (\amendexp, \statrev, \reqrev, \riskrev) that align with Japanese corporate legal practice. We analyzed leading LLMs using both human evaluation and automated evaluation. To clarify the outcomes of this paper, we revisit the research questions (RQ1–3) presented at the beginning and summarize the answers obtained from our research below:

\textbf{RQ1:} In \reqrev, which involves explicit revision instructions, all models demonstrated high performance. In contrast, for revision tasks with a higher degree of abstraction, such as \statrev~and \riskrev, unnecessary alterations and structural breakdowns occurred frequently, confirming that the nature of the task can expose the model's weaknesses.
\textbf{RQ2:} A comparison with the results of general-purpose benchmarks published by OpenAI revealed that performance degradation due to a reduction in model scale is far more severe in the legal tasks proposed in this study than in general-purpose tasks. The results demonstrate that scores on conventional tasks alone cannot predict practical performance in specialized domains, supporting the importance of domain-specific benchmarks.
\textbf{RQ3:} For tasks like \amendexp, which involve summarizing amendments and their impacts, a significant correlation was found between the two evaluation methods. For \statrev, \reqrev, and \riskrev, moderate correlations were also confirmed for certain aspects, with high agreement on criteria with linguistic grounding, such as the absence of unnecessary changes and coverage. However, the correlation was weak for aspects like document structural consistency and the naturalness of technical terminology, making a complete substitution difficult. 
Overall, our findings suggest that automated evaluation is practical for screening or as a supplementary tool in environments where experts are not available.

As shown above, the analysis for each research question has concretely clarified the strengths and limitations of LLMs. Based on this, our benchmark provides a common foundation for evaluating LLM performance on legal tasks by releasing the data design, evaluation design, and experimental setup for reproducibility.

\paragraph{Future Work}

We can consider several future directions for this research. First, since the proposed task framework is language-independent, it is expected to apply to other languages and legal systems, such as those in English-speaking countries or civil law jurisdictions. This attempt would enable a comparative analysis of how differences in legal systems affect the editing behavior of LLMs. Second, we plan to expand the dataset and explore its application for few-shot learning and fine-tuning models specialized for specific legal tasks. Finally, we can extend the tasks to include other legal documents besides contracts, such as complaints, legal opinions, and internal regulations, to evaluate the capabilities of LLMs in a more diverse range of legal practice scenarios.

\section{Limitations}

This study has several limitations. First, the evaluation was limited to three major closed-source LLMs and did not include the open-source models that have seen recent performance improvements. Validation with a broader range of models is a task for the future. Second, the dataset consists of 25 samples per task, totaling 100 samples, which may introduce a bias toward specific contract types or legal fields. To enhance statistical robustness, it is necessary to expand the diversity and scale of the dataset.

\nocite{*}
\section{References}

\bibliographystyle{lrec2026-natbib}
\bibliography{reference,languageresource}

\appendix

\section{Dataset Example}

In this section, we present concrete examples of the input data and gold data for each task.

\subsection{\amendexp}

As an example of \amendexp, we show a case related to an amendment to the Penal Code.
Information on legal amendments was obtained from the Digital Agency’s e-Gov Law Database.
The text before the amendment (version enforced on May 23, 2025) is available at \url{https://laws.e-gov.go.jp/law/140AC0000000045/20250523_507AC0000000039}
,
and the text after the amendment (version enforced on June 1, 2025) is available at \url{https://laws.e-gov.go.jp/law/140AC0000000045/20250601_504AC0000000067}
.
Based on these, an example output summarizing the overview of the amendment and its impact on contracts is shown in summary~\ref{box:full_output_amend_exp}.

\subsection{\statrev}

As an example of \statrev, we show a case related to an amendment to labor law.
The text before the amendment (version enforced on January 18, 2023) is available at \url{https://laws.e-gov.go.jp/law/322M40000100023/20230118_505M60000100006}, and the text after the amendment (version enforced on April 1, 2024) is available at \url{https://laws.e-gov.go.jp/law/322M40000100023/20240401_505M60000100039}.
Contract~\ref{box:full_example_stat_rev_pre_contract} corresponds to the pre-amendment contract, and contract~\ref{box:full_example_stat_rev_post_contract} corresponds to the post-amendment contract.

\subsection{\reqrev~and \riskrev}

As examples of \reqrev~and \riskrev, we present a case involving the revision of a service agreement.
In both \reqrev~and \riskrev~examples, the pre-edit contract corresponds to contract~\ref{box:full_example_pre_contract_req_risk_rev}, and the post-edit contract corresponds to contract~\ref{box:full_example_post_contract_req_risk_rev}.
Although the two tasks share a common input structure, the user instructions differ:
\reqrev~uses prompt~\ref{box:given_prompt_reqrev}, while \riskrev~uses prompt~\ref{box:given_prompt_riskrev}.

\section{Prompt settings}
Since each task was handled by different personnel, there are differences in implementation approaches.
\amendexp~and \statrev~directly call APIs, while \reqrev~and \riskrev~use LangChain.
Because the input/output formats differ among tasks, no direct comparison was made; this difference does not affect the analysis results.
For reproducibility, all prompt templates are publicly available.

\subsection{Inference prompts}

This subsection describes the inference-time prompt settings for each task. We used prompt~\ref{box:inference_prompt_amendexp} for \amendexp, prompt~\ref{box:inference_prompt_statrev} for \statrev, prompt~\ref{box:inference_prompt_reqrev} for \reqrev, and prompt~\ref{box:inference_prompt_riskrev} for \riskrev.

\subsection{Evaluation prompt}

This subsection describes the evaluation-time prompt settings for each task.
During evaluation, task-specific legal-domain evaluation criteria were applied to each task’s output, and the evaluation LLM was instructed to produce structured outputs in JSON format with a fixed schema.
As inputs, we provided \verb|{original_prompt}|, \verb|{instruction}|, \verb|{llm_response}|, and the ground-truth data \verb|{ground_truth}|, enabling automatic evaluation by the LLM.

The evaluation prompts used were: prompt~\ref{box:evaluation_prompt_amendexp} for \amendexp, prompt~\ref{box:evaluation_prompt_statrev} for \statrev, prompt~\ref{box:evaluation_prompt_reqrev} for \reqrev, and prompt~\ref{box:evaluation_prompt_riskrev} for \riskrev. We used LangChain to generate structured output.

\tiny
\onecolumn

\begin{summary}{An Example of the output for \amendexp}{full_output_amend_exp}
\# 改正の概要

刑事施設における受刑者の処遇及び執行猶予制度等のより一層の充実を図るため、懲役及び禁錮を廃止して拘禁刑を創設。\\

\# 改正による契約書への影響

ストックオプション割当契約等、個人に何らかの権利を付与する契約において、その個人が刑罰に処せられた場合はその権利を喪失する旨が定められている場合がある。その規定において、「懲役」「禁錮」という文言が使われていたら、「拘禁刑」に改正する必要がある。
\end{summary}

\begin{contract}{An example of pre-revised contract for \statrev}{full_example_stat_rev_pre_contract}
**雇用契約書兼労働条件通知書（正社員）**

株式会社AAA（以下「使用者」という。）とBBB（以下「従業員」という。）とは、以下のとおり雇用契約（以下「本契約」という。）を締結する。

1. **（目的）**

使用者は、従業員を正社員として雇用し、賃金を支払うことを約し、従業員は使用者の指揮に従い誠実に勤務することを約する。

2. **（雇用期間等）**  
1. 本契約は、雇用期間の定めのない契約とする。  
2. 従業員の就業開始日は、2025年4月1日とする。  
3. 就業開始日から3か月間は試用期間とする。

3. **（就業場所及び業務内容）**

従業員は、下記の就業場所において下記の業務を使用者の指示に従い誠実に行う。

記  
就業場所	　東京本社  
業　　務	　営業業務  
以上

4. **（就業時間）**  
1. 従業員の就業時間は、以下のとおりとする。  
   始業　午前9時  
   終業　午後18時  
2. 休憩時間は、60分とする。  
3. 使用者は、業務の必要がある場合には、法令の範囲内で本条第1項の各時刻を変更し、時間外労働を命ずることがある。

5. **（休日）**  
1. 従業員の休日は、以下のとおりとする。  
1) 土曜日  
2) 日曜日  
3) 年末年始（12月29日から12月31日まで並びに1月2日及び1月3日）  
4) 国民の祝日に関する法律に定める休日  
5) 国民の祝日が日曜日に当たるときはその翌日  
2. 使用者は業務の必要がある場合には、前項各号の休日に従業員を臨時に就業させ、他の日を振替休日とすることがある。

6. **（休暇）**  
1. 従業員は、法令に従い、所定の年次有給休暇を取得することができる。  
2. 従業員は、以下のとおり特別休暇を取得することができる。  
1) 有給　慶弔休暇、ボランティア休暇、リフレッシュ休暇  
2) 無給　私傷病休暇

7. **（賃金）**

従業員の賃金は、以下のとおりとする。

1) 従業員の賃金は、職能制とする。  
2) 基本給は、金350,000円とする。  
3) 諸手当として以下のものを支給する。  
1. 通勤手当	月額40,000 円まで／計算方法：距離に応じて支給  
2. 職務手当	月額100,000 円まで／計算方法：職務遂行能力に応じて支給  
4) 所定時間外労働等に対する割増賃金の計算方法は、以下のとおりとする。  
1. 所定時間外かつ法定時間内の労働にかかる部分　100%
2. 法定時間外　1か月60時間以下の労働にかかる部分　125%
   1か月60時間を超える労働にかかる部分　150%
3. 休日労働（法定休日に労働した場合）　135％  
4. 休日労働（所定休日（法定休日を除く。）に労働した場合）　100％  
5. 深夜労働（午後10時から午前5時までの間）　125％  
5) 賃金締切日は、毎月末日とする。  
6) 賃金支払日は、毎月25日とする（ただし、同日が金融機関の営業日でない場合はその直前営業日とする。）。  
7) 賃金は、全額を通貨にて支払うか、又は従業員指定の銀行口座に振り込む方法で支払う。なお、振込手数料は使用者の負担とする。  
8) 使用者の業績及び経営状態、従業員の評価等を考慮し、昇給する場合がある。

8. **（退職に関する事項）**

退職に関する事項は、以下のとおりとする。

1) 65歳をもって定年とし、定年に達した日の属する月の末日をもって退職とする。定年後も引き続き雇用されることを希望し、解雇事由又は退職事由に該当しない従業員については、満65歳までこれを継続雇用する。  
2) 自己都合で退職する場合には、退職希望日の1か月前までに申し出ることとする。  
3) 解雇については就業規則に記載のとおりとする。  
4) 退職金は支給しない。

9. **（遵守事項）**

従業員は、使用者に対し、以下の事項を遵守することを誓約する。

1) 法令、諸規則、諸規程又は業務命令等を遵守すること。  
2) 業務上の機密に属する事項（個人情報を含む。）を在職中はもとより退職後もこれを第三者に開示又は漏洩しないこと。  
3) 使用者の書面による承諾がない限り、在職中はもとより退職後2年間は、使用者と同一若しくは類似の事業を自ら営み、又は使用者と競合する事業を営む会社の役員若しくは従業員に就任、就職しないこと。  
4) 使用者の信用又は名誉を毀損する行為をしないこと。

10. **（社会保険等の適用）**

　従業員の社会保険等の適用は、以下のとおりとする。

1) 健康保険：適用する  
2) 厚生年金保険：適用する  
3) 雇用保険：適用する  
4) 中小企業退職金共済制度：加入している  
5) 企業年金制度：無し

11. **（その他）**  
1. 従業員の労働条件は、本契約に定めるもののほか、使用者の就業規則に定めるところによる。  
2. 本契約若しくは前項の各書類に定めのない事項又はこれらの解釈について疑義が生じたときは、使用者及び従業員の間で誠意をもって協議の上、解決する。

12. **（準拠法・合意管轄）**  
1. 本契約の準拠法は日本法とし、これに従って解釈される。  
2. 本契約に関する紛争については、東京地方裁判所を第一審の専属的合意管轄裁判所とする。

上記契約締結の証として本契約2通を作成し、使用者及び従業員双方署名又は記名押印の上、各1通を保有する。

年    月    日

（使用者）  
住　所  
会社名  
代表者

（従業員）  
住　所  
会社名  
代表者  

\end{contract}

\begin{contract}{An example of post-revised contract for \statrev}{full_example_stat_rev_post_contract}
**雇用契約書兼労働条件通知書（正社員）**

株式会社AAA（以下「使用者」という。）とBBB（以下「従業員」という。）とは、以下のとおり雇用契約（以下「本契約」という。）を締結する。

1. **（目的）**

使用者は、従業員を正社員として雇用し、賃金を支払うことを約し、従業員は使用者の指揮に従い誠実に勤務することを約する。

2. **（雇用期間等）**  
1. 本契約は、雇用期間の定めのない契約とする。  
2. 従業員の就業開始日は、2025年4月1日とする。  
3. 就業開始日から3か月間は試用期間とする。

3. **（就業場所及び業務内容）**

従業員は、下記の就業場所において下記の業務を使用者の指示に従い誠実に行う。

記  
就業場所	　雇入れ直後：東京本社　　変更の範囲：会社の別途指定する就業場所  
業　　務	　雇入れ直後：営業　　　　変更の範囲：会社の別途指定する業務  
以上

4. **（就業時間）**  
1. 従業員の就業時間は、以下のとおりとする。  
   始業　午前9時  
   終業　午後18時  
2. 休憩時間は、60分とする。  
3. 使用者は、業務の必要がある場合には、法令の範囲内で本条第1項の各時刻を変更し、時間外労働を命ずることがある。

5. **（休日）**  
1. 従業員の休日は、以下のとおりとする。  
1) 土曜日  
2) 日曜日  
3) 年末年始（12月29日から12月31日まで並びに1月2日及び1月3日）  
4) 国民の祝日に関する法律に定める休日  
5) 国民の祝日が日曜日に当たるときはその翌日  
2. 使用者は業務の必要がある場合には、前項各号の休日に従業員を臨時に就業させ、他の日を振替休日とすることがある。

6. **（休暇）**  
1. 従業員は、法令に従い、所定の年次有給休暇を取得することができる。  
2. 従業員は、以下のとおり特別休暇を取得することができる。  
1) 有給　慶弔休暇、ボランティア休暇、リフレッシュ休暇  
2) 無給　私傷病休暇

7. **（賃金）**

従業員の賃金は、以下のとおりとする。

1) 従業員の賃金は、職能制とする。  
2) 基本給は、金350,000円とする。  
3) 諸手当として以下のものを支給する。  
1. 通勤手当	月額40,000 円まで／計算方法：距離に応じて支給  
2. 職務手当	月額100,000 円まで／計算方法：職務遂行能力に応じて支給  
4) 所定時間外労働等に対する割増賃金の計算方法は、以下のとおりとする。  
1. 所定時間外かつ法定時間内の労働にかかる部分　100%
2. 法定時間外　1か月60時間以下の労働にかかる部分　125%
   1か月60時間を超える労働にかかる部分　150%
3. 休日労働（法定休日に労働した場合）　135％  
4. 休日労働（所定休日（法定休日を除く。）に労働した場合）　100％  
5. 深夜労働（午後10時から午前5時までの間）　125％  
5) 賃金締切日は、毎月末日とする。  
6) 賃金支払日は、毎月25日とする（ただし、同日が金融機関の営業日でない場合はその直前営業日とする。）。  
7) 賃金は、全額を通貨にて支払うか、又は従業員指定の銀行口座に振り込む方法で支払う。なお、振込手数料は使用者の負担とする。  
8) 使用者の業績及び経営状態、従業員の評価等を考慮し、昇給する場合がある。

8. **（退職に関する事項）**

退職に関する事項は、以下のとおりとする。

1) 65歳をもって定年とし、定年に達した日の属する月の末日をもって退職とする。定年後も引き続き雇用されることを希望し、解雇事由又は退職事由に該当しない従業員については、満65歳までこれを継続雇用する。  
2) 自己都合で退職する場合には、退職希望日の1か月前までに申し出ることとする。  
3) 解雇については就業規則に記載のとおりとする。  
4) 退職金は支給しない。

9. **（遵守事項）**

従業員は、使用者に対し、以下の事項を遵守することを誓約する。

1) 法令、諸規則、諸規程又は業務命令等を遵守すること。  
2) 業務上の機密に属する事項（個人情報を含む。）を在職中はもとより退職後もこれを第三者に開示又は漏洩しないこと。  
3) 使用者の書面による承諾がない限り、在職中はもとより退職後2年間は、使用者と同一若しくは類似の事業を自ら営み、又は使用者と競合する事業を営む会社の役員若しくは従業員に就任、就職しないこと。  
4) 使用者の信用又は名誉を毀損する行為をしないこと。

10. **（社会保険等の適用）**

　従業員の社会保険等の適用は、以下のとおりとする。

1) 健康保険：適用する  
2) 厚生年金保険：適用する  
3) 雇用保険：適用する  
4) 中小企業退職金共済制度：加入している  
5) 企業年金制度：無し

11. **（その他）**  
1. 従業員の労働条件は、本契約に定めるもののほか、使用者の就業規則に定めるところによる。  
2. 本契約若しくは前項の各書類に定めのない事項又はこれらの解釈について疑義が生じたときは、使用者及び従業員の間で誠意をもって協議の上、解決する。

12. **（準拠法・合意管轄）**  
1. 本契約の準拠法は日本法とし、これに従って解釈される。  
2. 本契約に関する紛争については、東京地方裁判所を第一審の専属的合意管轄裁判所とする。

上記契約締結の証として本契約2通を作成し、使用者及び従業員双方署名又は記名押印の上、各1通を保有する。

年    月    日

（使用者）  
住　所  
会社名  
代表者

（従業員）  
住　所  
会社名  
代表者  

\end{contract}

\begin{prompt}{Inference Prompt for \reqrev}{given_prompt_reqrev}
・委託者の立場から、以下の指示に従って契約書の内容を確認・修正してください。\\
\phantom{XX}・以下の内容について、適切な表現で追加又は修正してください。\\
\phantom{XXXX} - 「成果物が、第三者の権利を侵害しない」旨を、受託者が保証する規定を、第12条（納入物の知的財産権等）と第13条（知的財産権の侵害）の間に追加\\
\phantom{XXXX} - 第７条（資料等の提供）に「資料の複製を禁止する旨」を追加\\
\phantom{XXXX} - 第８条（再委託）を「委託者の事前の書面による承諾を得た場合に限り、再委託可能とする」旨に修正\\
\phantom{XXXX} - 第１０条（契約不適合責任）を「委託者が指定した方法により履行の追完を行う」旨、「委託者が、履行の追完請求を行うことなく、委託料の減額を請求することができる」旨、「契約不適合責任を追及できる期間を委託者が契約不適合を知った時から1年以内」に修正\\
\phantom{XX} ・条項の追加・削除を行う場合は、条番号のずれや他条項での引用の整合性も必ず修正してください。\\
\end{prompt}

\begin{prompt}{Inference Prompt for \riskrev}{given_prompt_riskrev}
・委託者の立場から、以下の指示に従って契約書の内容を確認・修正してください。\\
 \phantom{XX}   ・以下の内容について、契約書内に該当する文言がない場合は、適切な表現を追加してください。\\
 \phantom{XXXX}       - 成果物が第三者の権利を侵害しない旨\\
   \phantom{XXXX}       - 資料の複製を禁止する旨\\
  \phantom{XX}    ・以下の内容について、契約書内に該当するテーマや項目が含まれている場合は、当該表現を委託側に有利に修正してください。削除することが有利になる場合は\\
  \phantom{XXXX}、削除も可能とします。\\
  \phantom{XXXX}        - 再委託に関する条項\\
   \phantom{XXXX}       - 契約不適合責任に関する内容\\
  \phantom{XXXX}        - 連帯保証人に関する内容\\
   \phantom{XXXX}       - 人材紹介の手数料に関する内容\\
   \phantom{XX}   ・条項の追加・削除を行う場合は、条番号のずれや他条項での引用の整合性も必ず修正してください。\\
\end{prompt}

\onecolumn
\begin{contract}{An example pre-revised contract for \reqrev~and \riskrev}{full_example_pre_contract_req_risk_rev}
**業務委託契約書**

委託者〇〇（以下「甲」という。）及び受託者○○（以下「乙」という。）は、契約要綱記載の業務の委託に関して、次のとおり、業務委託契約（以下「本契約」という。）を締結する。\\

**＜契約要綱＞**\\

| 本委託業務の表示 |   |\\
| :---- | ----- |\\
| 納入物 |   |\\
| 納期・検査完了期日 | 納期：〇年○月○日 検査完了期日：納入後○日以内 |\\
| 納入場所 |   |\\
| 業務委託料 | 金　〇　円　（消費税抜金額〇円　消費税額〇円） |\\
| 業務委託料の支払時期 | 第9条に定める【納品／検査】が完了した日から○日以内 |\\
| 支払方法 | 【全額現金払（乙の指定する銀行口座への口座振込による。なお、支払期日が金融機関の休業日に当たる場合、【前／翌】営業日に支払う。また、振込手数料は甲の負担とする。】 |\\
| 特約事項 |   |\\

第1条（業務の委託）

1.    甲は、契約要綱記載の業務（以下「本委託業務」という。）を乙に委託し、乙はこれを受託する。

2.    乙は、本委託業務を、善良なる管理者の注意義務をもって遂行し、本委託業務に関連する法令を遵守し、本契約にて定められた業務の遂行について、事業主として財政上及び法律上を含む全ての責任を負う。

第2条（本委託業務の詳細）

本委託業務の詳細については、甲乙別途協議のうえ定めるものとし、必要に応じて手順書等を作成する。

第3条（業務委託料及びその支払方法）

1.	甲は乙に対し、本委託業務の対価として、契約要綱記載の業務委託料並びにこれに対する消費税及び地方消費税（以下、併せて「本業務委託料」という。）を支払う。

2.	乙は、第9条に定める検査が完了した場合、甲に対し本業務委託料にかかる請求書を交付するものとし、甲は、本業務委託料を契約要綱記載の期日までに、契約要綱記載の方法により支払う。

第4条（納期）

本委託業務にかかる納期は、契約要綱に定めるとおりとする。

第5条（責任者）

1.    甲及び乙は、本契約締結後速やかに、本契約における各自の責任者をそれぞれ選任し、互いに書面により、相手方に通知する。

2.	甲及び乙は、事前に書面により相手方に通知することにより、責任者を変更することができる。

3.    甲及び乙の責任者は、本委託業務遂行に必要な意思決定、指示、同意等を行う権限及び責任を有する。本委託業務に関する要請、指示、依頼その他の連絡、確認等は原則として責任者を通じて行わなければならない。

第6条（業務従事者）

1.    乙は、労働法規その他関係法令に基づき、本委託業務に従事する乙の従業員（以下「業務従事者」という。）に対し雇用主としての一切の義務を負うものとし、業務従事者に対する本委託業務遂行に関する指示、労務管理、安全衛生管理等に関する一切の指揮命令を行う。

2.    乙は、本委託業務遂行上、業務従事者が甲の事務所等に立ち入る場合、甲の防犯、秩序維持等に関する諸規則を当該業務従事者に遵守させなければならない。

第7条（資料等の提供）

1.	甲は乙に対し、本委託業務遂行に必要な資料等を提供する。乙は甲から提供された本委託業務に関する資料等を善良な管理者の注意をもって管理、保管するものとし、本委託業務以外の用途に使用してはならない。

2.    本委託業務遂行上、甲の事務所等で乙が作業を実施する必要がある場合、甲は当該作業実施場所（当該作業実施場所における必要な機器、設備等作業環境を含む。）を、甲乙協議の上、乙に提供するものとする。

3.    甲が前各項により乙に提供する資料等又は作業実施場所に関し、内容等の誤り又は甲の提供遅延があった場合、これにより生じた乙の本委託業務の履行遅滞、納入物の契約不適合等の結果については、乙はその責を免れる。

4.    甲から提供を受けた資料等が本委託業務遂行上不要となったときは、乙は遅滞なく資料等につき甲の指示に従い返還等の処置を行うものとする。

第8条（再委託）

1.	乙は、乙の責任において、本委託業務の一部を第三者に再委託することができる。ただし、乙は、甲が要請した場合、再委託先の名称及び住所等を甲に報告しなければならない。

2.    甲が前項に定める再委託先が不適切であると合理的な理由により判断した場合、甲は、乙に、書面によりその理由を通知し、当該第三者に対する再委託の中止を請求することができる。

3.    乙は、本契約に基づいて乙が甲に対して負担するのと同様の義務を、再委託先に負わせなければならず、また、本委託業務に関する再委託先の行為について、自らの行為と同様の責任を負うものとする。ただし、甲の指定した再委託先の行為については、乙に故意又は重過失がある場合を除き、責任を負わない。

第9条（納入及び検収）

1.    乙は、甲に対し、契約要綱記載の納期までに、契約要綱記載の納入物を納入する。乙が納入に際し、甲に対して必要な協力を要請した場合、甲は、すみやかにこれに応じなければならない。

2.    甲は、契約要綱記載の検査完了期日までに、甲乙が合意した内容の検査を行い、納入物が当該検査に合格した場合、乙に対し検査合格証を交付する。

3.    甲は、納入物が前項の検査に合格しない場合、乙に対し不合格となった具体的な理由を明示した書面を交付して修正又は追完を求めることができ、乙は、協議の上定めた期限内に納入物を無償で修正又は追完したうえで甲に再度納入し、検査を受けなければならない。

4.    検査完了期日までに前2項に定める検査合格証又は不合格の理由を明示した書面が乙に対し交付されない場合、甲が検査遅延の理由について書面で合理的な理由を明示したときを除き、納入物は、本条所定の検査に合格したものとみなす。

5.    本条所定の検査に合格したことをもって、納入物の検収完了とし、納入物の引渡しが完了したものとみなす。

第10条（契約不適合責任）

1.	納入物に本契約の内容との不適合があった場合、甲は、乙に対し、当該納入物の無償での修補、代替品の納入又は不足分の納入等の方法による履行の追完を請求することができる。ただし、乙は、甲に不相当な負担を課すものでないときは、甲が請求した方法と異なる方法による履行の追完をすることができる。

2.	甲が、前項により相当の期間を定めて履行の追完の催告をし、その期間内に履行の追完がないときは、甲は、その不適合の程度に応じて業務委託料の減額を請求することができる。

3.    乙が本契約の内容との不適合のある納入物を甲に引き渡した場合において、甲が引渡した時から【1】年以内にその旨を乙に通知しないときは、甲は、その不適合を理由として、履行の追完の請求、業務委託料の減額の請求、損害賠償の請求及び本契約の解除をすることができない。

4.    甲の責めに帰すべき事由により契約不適合が生じたときは、甲は、履行の追完、業務委託料の減額又は返還、損害賠償の請求及び契約の解除をすることができない。

第11条（納入物の所有権）

1.    納入物の所有権は、第9条に定める検収が完了したときに、乙から甲へ移転する。

2.    納入物の滅失、毀損等の危険負担は、納入前については乙が、納入後については甲がそれぞれ負担する。

第12条（納入物の知的財産権等）

1.	納入物に関する著作権（著作権法第27条及び第28条に規定する権利を含む。）は、当該納入物にかかる本業務委託料の支払いが完了した時点で、甲に帰属する。ただし、乙又は第三者が従前から保有していた著作権は、乙又は当該第三者に留保されるものとする。

2.	乙は、納入物について著作者人格権を行使しないものとする。

3.	本委託業務遂行の過程で生じた発明その他の知的財産又はノウハウ等（以下、併せて「発明等」という。）に係る特許権その他の知的財産権（特許その他の知的財産権を受ける権利を含む。）、ノウハウ等に関する権利（以下、特許権その他の知的財産権、ノウハウ等に関する権利を総称して「特許権等」という。）は、第1項に基づき甲に帰属する著作権を除き、当該発明等を行った者が属する当事者に帰属する。

4.    乙は、前項に基づき特許権等を保有する場合、甲に対し、甲が納入物を使用するのに必要な範囲において、当該特許権等の通常実施権を許諾する。なお、当該許諾の対価は、本業務委託料に含まれる。

第13条（知的財産権の侵害）

1.    甲は、乙による本委託業務の履行に起因して第三者から知的財産権の侵害の申立てを受けたときは、速やかに乙に対して申立ての事実及び内容を通知する。

2.    前項の場合において、乙は、甲が第三者との交渉又は訴訟の遂行に関し、乙に実質的な参加の機会及び決定の権限を与え、必要な援助を行ったときは、甲が支払うべきとされた損害賠償額を負担する。ただし、当該知的財産権の侵害が専ら甲の責に帰すべき事由により生じた場合には、乙は賠償する義務を負わない。

3.    乙は、本委託業務の履行に起因する第三者の知的財産権の侵害に関し、本条の定めに従った責任のみを負い、それ以外に一切責任を負わない。

第14条（秘密保持）

1.	甲及び乙は、本契約の存在及び内容並びに本契約に関連して知った相手方に関する情報（以下「秘密情報」という。）を、相手方の書面による事前の承諾なく第三者に対して開示、提供又は漏洩してはならない。ただし、次のいずれかに該当することを証明できる情報については、秘密情報から除くものとする。

(1)   提供又は開示を受けた際、既に自己が保有していた情報

(2)   提供又は開示を受けた際、既に公知となっている情報

(3)   提供又は開示を受けた後、自己の責めによらずに公知となった情報

(4)   正当な権限を有する第三者から秘密保持義務を負わずに適法に取得した情報

(5)   秘密情報によることなく独自に開発又は取得した情報

(6)   法令の規定又は権限ある官公庁からの開示の要求に基づき開示しなければならない情報

2.    甲及び乙は、秘密情報を本契約の目的以外に使用してはならない。

3.    甲及び乙は、秘密情報を、本契約の遂行のために知る必要のある自己の役員及び従業員（以下「役員等」という。）並びに弁護士等の法令上守秘義務を負う専門家に限り開示等することができる。この場合、本条に基づき秘密情報の受領者が負担する義務と同等の義務を、開示等を受けた当該役員等に退職後も含め課さなければならない。

4.    本契約が終了した場合、甲及び乙は、相手方の指示に従って、秘密情報を返還し、又は、破棄するものとする。なお、相手方は受領者に対し、秘密情報等の返還又は破棄を証明する文書の提出を求めることができる。

第15条（個人情報）

1.    乙は、本委託業務遂行に際して甲より取扱いを委託された個人情報（個人情報の保護に関する法律に定める個人情報をいう。以下同じ。）を適切に管理し、他に開示、漏洩、又は公開してはならない。

2.    乙は、個人情報を、本契約の目的の範囲内でのみ使用しなければならない。

3.	個人情報が本委託業務遂行上不要となったときは、乙は遅滞なく甲の指示に従い返還等の処置を行うものとする。

第16条（報告等）

1.    甲が、本委託業務の遂行状況又は秘密情報若しくは個人情報の管理状況について乙に報告を求めたときは、乙は、これを速やかに甲に報告しなければならない。

2.    本委託業務に関する問題が発生し、若しくはそのおそれがある場合又は乙に本契約違反があると合理的理由により認められる場合には、甲は、乙に検査日の○日前までに書面で通知することにより、乙の営業時間中、乙の事業所に立ち入り、本契約の履行状況を検査することができる。

第17条（損害賠償）

甲又は乙は、自己の責めに帰すべき事由により本契約に違反して相手方に損害を与えた場合、その損害を賠償する義務を負う。

第18条（遅延損害金）

甲及び乙は、本契約に基づく債務の支払を遅滞したときは、相手方に対し、当該債務の支払期日の翌日から支払済みまで年【14.6】％（年365日の日割計算）の割合による遅延損害金を支払わなければならない。

第19条（解除等）

1.    甲又は乙が本契約のいずれかの条項に違反し、相手方からその是正を要求する通知を受領した後○日以内にその違反を是正しない場合は、相手方は、違反者にその旨を通知することにより、本契約を直ちに解除することができる。

2.    甲は、乙が以下の各号に定める事由のいずれかに該当した場合には、何らの催告なくして、直ちに本契約を解除することができる。

(1)  重大な過失又は背信行為があった場合

(2)  手形若しくは小切手の不渡りが生じたとき、電子交換所の取引停止処分を受けたとき、又は支払停止の状態に陥ったとき。

(3)  第三者より差押え、仮差押え、仮処分、強制執行などを受けたとき、又は公租公課の滞納処分を受けたとき。

(4)  破産手続、特別清算、民事再生手続又は会社更生手続開始の申立てがあったとき。

(5)  債務整理の通知がなされたとき

(6)  その他財務状態が著しく悪化し、又は悪化のおそれがあるとき。

(7)  その他前各号に準ずるような本契約を継続し難い重大な事由が発生した場合

第20条（契約終了時の措置）

本契約が事由の如何を問わず本委託業務の履行完了前に終了した場合、甲は、乙に対し、既に履行された本委託業務の内容に応じた本業務委託料を支払わなければならない。なお、本条の規定は、第17条並びに民法641条及び642条の規定を排除するものではない。

第21条（反社会的勢力の排除）

1.    甲及び乙は、自社、自社の株主・役員その他自社を実質的に所有し、若しくは支配するものが、現在、暴力団、暴力団員、暴力団員でなくなった時から5年を経過しない者、暴力団準構成員、暴力団関係企業、総会屋等、社会運動等標ぼうゴロ又は特殊知能暴力集団等、その他これらに準ずる者（以下、これらを「暴力団員等」という。）に該当しないこと、及び次の各号のいずれにも該当しないことを表明し、かつ将来にわたっても該当しないことを確約する。

(1)  暴力団員等が経営を支配していると認められる関係を有すること

(2)  暴力団員等が経営に実質的に関与していると認められる関係を有すること

(3)  自己、自社若しくは第三者の不正の利益を図る目的又は第三者に損害を加える目的をもってする等、不当に暴力団員等を利用していると認められる関係を有すること

(4)  暴力団員等に対して資金等を提供し、又は便宜を供与する等の関与をしていると認められる関係を有すること

(5)  役員又は経営に実質的に関与している者が暴力団員等と社会的に非難されるべき関係を有すること

2.    甲及び乙は、暴力団員等と取引関係を有してはならず、事後的に、暴力団員等との取引関係が判明した場合には、これを相当期間内に解消できるよう必要な措置を講じる。

3.    甲及び乙は、相手方が本条の表明又は確約に違反した場合、何らの通知又は催告をすることなく直ちに本契約の全部又は一部について、履行を停止し、又は解除することができる。この場合において、表明又は確約に違反した当事者は、相手方の履行停止又は解除によって被った損害の賠償を請求することはできない。

4.    甲及び乙は、相手方が本条の表明又は確約に違反した場合、これによって被った一切の損害の賠償を請求することができる。

第22条（本契約上の地位譲渡等の禁止）

甲及び乙は、相手方の書面による事前の承諾がない限り、本契約上の権利及び義務並びに本契約上の地位を第三者へ譲渡し、又は担保に供してはならない。

第23条（存続条項）

本契約が終了した場合でも、第10条から第15条、第17条、第18条、第20条、第21条3項から4項、及び、本条から第26条は有効に存続する。ただし、第14条については、本契約の終了後【2】年間に限り、その効力を有する。

第24条（準拠法）

本契約は、日本法を準拠法とする。

第25条（合意管轄）

本契約に関する一切の紛争については、法令に専属管轄の定めがある場合を除き、被告の住所地を管轄する地方裁判所又は簡易裁判所を第一審の専属的合意管轄裁判所とする。

第26条（協議）

甲及び乙は、本契約に定めがない場合及び本契約の条項の解釈について疑義が生じた場合は、民法その他の法令及び慣行に従い、誠意をもって協議し、解決するものとする。

本契約の成立を証するため本書2通を作成し、各自記名押印のうえ、各1通を保有する。

〇年〇月〇日

　　　　　　　　委託者（甲）

　　　　　　　　受託者（乙）

\end{contract}

\begin{contract}{An example post-revised contract for \reqrev~and \riskrev}{full_example_post_contract_req_risk_rev}
**業務委託契約書**

委託者〇〇（以下「甲」という。）及び受託者○○（以下「乙」という。）は、契約要綱記載の業務の委託に関して、次のとおり、業務委託契約（以下「本契約」という。）を締結する。\\

**＜契約要綱＞**\\

| 本委託業務の表示 |   |\\
| :---- | ----- |\\
| 納入物 |   |\\
| 納期・検査完了期日 | 納期：〇年○月○日 検査完了期日：納入後○日以内 |\\
| 納入場所 |   |\\
| 業務委託料 | 金　〇　円　（消費税抜金額〇円　消費税額〇円） |\\
| 業務委託料の支払時期 | 第9条に定める【納品／検査】が完了した日から○日以内 |\\
| 支払方法 | 【全額現金払（乙の指定する銀行口座への口座振込による。なお、支払期日が金融機関の休業日に当たる場合、【前／翌】営業日に支払う。また、振込手数料は甲の負担とする。】 |\\
| 特約事項 |   |\\

第1条（業務の委託）

1.    甲は、契約要綱記載の業務（以下「本委託業務」という。）を乙に委託し、乙はこれを受託する。

2.    乙は、本委託業務を、善良なる管理者の注意義務をもって遂行し、本委託業務に関連する法令を遵守し、本契約にて定められた業務の遂行について、事業主として財政上及び法律上を含む全ての責任を負う。

第2条（本委託業務の詳細）

本委託業務の詳細については、甲乙別途協議のうえ定めるものとし、必要に応じて手順書等を作成する。

第3条（業務委託料及びその支払方法）

1.	甲は乙に対し、本委託業務の対価として、契約要綱記載の業務委託料並びにこれに対する消費税及び地方消費税（以下、併せて「本業務委託料」という。）を支払う。

2.	乙は、第9条に定める検査が完了した場合、甲に対し本業務委託料にかかる請求書を交付するものとし、甲は、本業務委託料を契約要綱記載の期日までに、契約要綱記載の方法により支払う。

第4条（納期）

本委託業務にかかる納期は、契約要綱に定めるとおりとする。

第5条（責任者）

1.    甲及び乙は、本契約締結後速やかに、本契約における各自の責任者をそれぞれ選任し、互いに書面により、相手方に通知する。

2.	甲及び乙は、事前に書面により相手方に通知することにより、責任者を変更することができる。

3.    甲及び乙の責任者は、本委託業務遂行に必要な意思決定、指示、同意等を行う権限及び責任を有する。本委託業務に関する要請、指示、依頼その他の連絡、確認等は原則として責任者を通じて行わなければならない。

第6条（業務従事者）

1.    乙は、労働法規その他関係法令に基づき、本委託業務に従事する乙の従業員（以下「業務従事者」という。）に対し雇用主としての一切の義務を負うものとし、業務従事者に対する本委託業務遂行に関する指示、労務管理、安全衛生管理等に関する一切の指揮命令を行う。

2.    乙は、本委託業務遂行上、業務従事者が甲の事務所等に立ち入る場合、甲の防犯、秩序維持等に関する諸規則を当該業務従事者に遵守させなければならない。

第7条（資料等の提供）

1.	甲は乙に対し、本委託業務遂行に必要な資料等を提供する。乙は甲から提供された本委託業務に関する資料等を善良な管理者の注意をもって管理、保管するものとし、本委託業務以外の用途に使用してはならない。

2.    本委託業務遂行上、甲の事務所等で乙が作業を実施する必要がある場合、甲は当該作業実施場所（当該作業実施場所における必要な機器、設備等作業環境を含む。）を、甲乙協議の上、乙に提供するものとする。

3.    甲が前各項により乙に提供する資料等又は作業実施場所に関し、内容等の誤り又は甲の提供遅延があった場合、これにより生じた乙の本委託業務の履行遅滞、納入物の契約不適合等の結果については、乙はその責を免れる。

4.    甲から提供を受けた資料等が本委託業務遂行上不要となったときは、乙は遅滞なく資料等につき甲の指示に従い返還等の処置を行うものとする。

第8条（再委託）

1.	乙は、乙の責任において、本委託業務の一部を第三者に再委託することができる。ただし、乙は、甲が要請した場合、再委託先の名称及び住所等を甲に報告しなければならない。

2.    甲が前項に定める再委託先が不適切であると合理的な理由により判断した場合、甲は、乙に、書面によりその理由を通知し、当該第三者に対する再委託の中止を請求することができる。

3.    乙は、本契約に基づいて乙が甲に対して負担するのと同様の義務を、再委託先に負わせなければならず、また、本委託業務に関する再委託先の行為について、自らの行為と同様の責任を負うものとする。ただし、甲の指定した再委託先の行為については、乙に故意又は重過失がある場合を除き、責任を負わない。

第9条（納入及び検収）

1.    乙は、甲に対し、契約要綱記載の納期までに、契約要綱記載の納入物を納入する。乙が納入に際し、甲に対して必要な協力を要請した場合、甲は、すみやかにこれに応じなければならない。

2.    甲は、契約要綱記載の検査完了期日までに、甲乙が合意した内容の検査を行い、納入物が当該検査に合格した場合、乙に対し検査合格証を交付する。

3.    甲は、納入物が前項の検査に合格しない場合、乙に対し不合格となった具体的な理由を明示した書面を交付して修正又は追完を求めることができ、乙は、協議の上定めた期限内に納入物を無償で修正又は追完したうえで甲に再度納入し、検査を受けなければならない。

4.    検査完了期日までに前2項に定める検査合格証又は不合格の理由を明示した書面が乙に対し交付されない場合、甲が検査遅延の理由について書面で合理的な理由を明示したときを除き、納入物は、本条所定の検査に合格したものとみなす。

5.    本条所定の検査に合格したことをもって、納入物の検収完了とし、納入物の引渡しが完了したものとみなす。

第10条（契約不適合責任）

1.	納入物に本契約の内容との不適合があった場合、甲は、乙に対し、当該納入物の無償での修補、代替品の納入又は不足分の納入等の方法による履行の追完を請求することができる。ただし、乙は、甲に不相当な負担を課すものでないときは、甲が請求した方法と異なる方法による履行の追完をすることができる。

2.	甲が、前項により相当の期間を定めて履行の追完の催告をし、その期間内に履行の追完がないときは、甲は、その不適合の程度に応じて業務委託料の減額を請求することができる。

3.    乙が本契約の内容との不適合のある納入物を甲に引き渡した場合において、甲が引渡した時から【1】年以内にその旨を乙に通知しないときは、甲は、その不適合を理由として、履行の追完の請求、業務委託料の減額の請求、損害賠償の請求及び本契約の解除をすることができない。

4.    甲の責めに帰すべき事由により契約不適合が生じたときは、甲は、履行の追完、業務委託料の減額又は返還、損害賠償の請求及び契約の解除をすることができない。

第11条（納入物の所有権）

1.    納入物の所有権は、第9条に定める検収が完了したときに、乙から甲へ移転する。

2.    納入物の滅失、毀損等の危険負担は、納入前については乙が、納入後については甲がそれぞれ負担する。

第12条（納入物の知的財産権等）

1.	納入物に関する著作権（著作権法第27条及び第28条に規定する権利を含む。）は、当該納入物にかかる本業務委託料の支払いが完了した時点で、甲に帰属する。ただし、乙又は第三者が従前から保有していた著作権は、乙又は当該第三者に留保されるものとする。

2.	乙は、納入物について著作者人格権を行使しないものとする。

3.	本委託業務遂行の過程で生じた発明その他の知的財産又はノウハウ等（以下、併せて「発明等」という。）に係る特許権その他の知的財産権（特許その他の知的財産権を受ける権利を含む。）、ノウハウ等に関する権利（以下、特許権その他の知的財産権、ノウハウ等に関する権利を総称して「特許権等」という。）は、第1項に基づき甲に帰属する著作権を除き、当該発明等を行った者が属する当事者に帰属する。

4.    乙は、前項に基づき特許権等を保有する場合、甲に対し、甲が納入物を使用するのに必要な範囲において、当該特許権等の通常実施権を許諾する。なお、当該許諾の対価は、本業務委託料に含まれる。

第13条（知的財産権の侵害）

1.    甲は、乙による本委託業務の履行に起因して第三者から知的財産権の侵害の申立てを受けたときは、速やかに乙に対して申立ての事実及び内容を通知する。

2.    前項の場合において、乙は、甲が第三者との交渉又は訴訟の遂行に関し、乙に実質的な参加の機会及び決定の権限を与え、必要な援助を行ったときは、甲が支払うべきとされた損害賠償額を負担する。ただし、当該知的財産権の侵害が専ら甲の責に帰すべき事由により生じた場合には、乙は賠償する義務を負わない。

3.    乙は、本委託業務の履行に起因する第三者の知的財産権の侵害に関し、本条の定めに従った責任のみを負い、それ以外に一切責任を負わない。

第14条（秘密保持）

1.	甲及び乙は、本契約の存在及び内容並びに本契約に関連して知った相手方に関する情報（以下「秘密情報」という。）を、相手方の書面による事前の承諾なく第三者に対して開示、提供又は漏洩してはならない。ただし、次のいずれかに該当することを証明できる情報については、秘密情報から除くものとする。

(1)   提供又は開示を受けた際、既に自己が保有していた情報

(2)   提供又は開示を受けた際、既に公知となっている情報

(3)   提供又は開示を受けた後、自己の責めによらずに公知となった情報

(4)   正当な権限を有する第三者から秘密保持義務を負わずに適法に取得した情報

(5)   秘密情報によることなく独自に開発又は取得した情報

(6)   法令の規定又は権限ある官公庁からの開示の要求に基づき開示しなければならない情報

2.    甲及び乙は、秘密情報を本契約の目的以外に使用してはならない。

3.    甲及び乙は、秘密情報を、本契約の遂行のために知る必要のある自己の役員及び従業員（以下「役員等」という。）並びに弁護士等の法令上守秘義務を負う専門家に限り開示等することができる。この場合、本条に基づき秘密情報の受領者が負担する義務と同等の義務を、開示等を受けた当該役員等に退職後も含め課さなければならない。

4.    本契約が終了した場合、甲及び乙は、相手方の指示に従って、秘密情報を返還し、又は、破棄するものとする。なお、相手方は受領者に対し、秘密情報等の返還又は破棄を証明する文書の提出を求めることができる。

第15条（個人情報）

1.    乙は、本委託業務遂行に際して甲より取扱いを委託された個人情報（個人情報の保護に関する法律に定める個人情報をいう。以下同じ。）を適切に管理し、他に開示、漏洩、又は公開してはならない。

2.    乙は、個人情報を、本契約の目的の範囲内でのみ使用しなければならない。

3.	個人情報が本委託業務遂行上不要となったときは、乙は遅滞なく甲の指示に従い返還等の処置を行うものとする。

第16条（報告等）

1.    甲が、本委託業務の遂行状況又は秘密情報若しくは個人情報の管理状況について乙に報告を求めたときは、乙は、これを速やかに甲に報告しなければならない。

2.    本委託業務に関する問題が発生し、若しくはそのおそれがある場合又は乙に本契約違反があると合理的理由により認められる場合には、甲は、乙に検査日の○日前までに書面で通知することにより、乙の営業時間中、乙の事業所に立ち入り、本契約の履行状況を検査することができる。

第17条（損害賠償）

甲又は乙は、自己の責めに帰すべき事由により本契約に違反して相手方に損害を与えた場合、その損害を賠償する義務を負う。

第18条（遅延損害金）

甲及び乙は、本契約に基づく債務の支払を遅滞したときは、相手方に対し、当該債務の支払期日の翌日から支払済みまで年【14.6】％（年365日の日割計算）の割合による遅延損害金を支払わなければならない。

第19条（解除等）

1.    甲又は乙が本契約のいずれかの条項に違反し、相手方からその是正を要求する通知を受領した後○日以内にその違反を是正しない場合は、相手方は、違反者にその旨を通知することにより、本契約を直ちに解除することができる。

2.    甲は、乙が以下の各号に定める事由のいずれかに該当した場合には、何らの催告なくして、直ちに本契約を解除することができる。

(1)  重大な過失又は背信行為があった場合

(2)  手形若しくは小切手の不渡りが生じたとき、電子交換所の取引停止処分を受けたとき、又は支払停止の状態に陥ったとき。

(3)  第三者より差押え、仮差押え、仮処分、強制執行などを受けたとき、又は公租公課の滞納処分を受けたとき。

(4)  破産手続、特別清算、民事再生手続又は会社更生手続開始の申立てがあったとき。

(5)  債務整理の通知がなされたとき

(6)  その他財務状態が著しく悪化し、又は悪化のおそれがあるとき。

(7)  その他前各号に準ずるような本契約を継続し難い重大な事由が発生した場合

第20条（契約終了時の措置）

本契約が事由の如何を問わず本委託業務の履行完了前に終了した場合、甲は、乙に対し、既に履行された本委託業務の内容に応じた本業務委託料を支払わなければならない。なお、本条の規定は、第17条並びに民法641条及び642条の規定を排除するものではない。

第21条（反社会的勢力の排除）

1.    甲及び乙は、自社、自社の株主・役員その他自社を実質的に所有し、若しくは支配するものが、現在、暴力団、暴力団員、暴力団員でなくなった時から5年を経過しない者、暴力団準構成員、暴力団関係企業、総会屋等、社会運動等標ぼうゴロ又は特殊知能暴力集団等、その他これらに準ずる者（以下、これらを「暴力団員等」という。）に該当しないこと、及び次の各号のいずれにも該当しないことを表明し、かつ将来にわたっても該当しないことを確約する。

(1)  暴力団員等が経営を支配していると認められる関係を有すること

(2)  暴力団員等が経営に実質的に関与していると認められる関係を有すること

(3)  自己、自社若しくは第三者の不正の利益を図る目的又は第三者に損害を加える目的をもってする等、不当に暴力団員等を利用していると認められる関係を有すること

(4)  暴力団員等に対して資金等を提供し、又は便宜を供与する等の関与をしていると認められる関係を有すること

(5)  役員又は経営に実質的に関与している者が暴力団員等と社会的に非難されるべき関係を有すること

2.    甲及び乙は、暴力団員等と取引関係を有してはならず、事後的に、暴力団員等との取引関係が判明した場合には、これを相当期間内に解消できるよう必要な措置を講じる。

3.    甲及び乙は、相手方が本条の表明又は確約に違反した場合、何らの通知又は催告をすることなく直ちに本契約の全部又は一部について、履行を停止し、又は解除することができる。この場合において、表明又は確約に違反した当事者は、相手方の履行停止又は解除によって被った損害の賠償を請求することはできない。

4.    甲及び乙は、相手方が本条の表明又は確約に違反した場合、これによって被った一切の損害の賠償を請求することができる。

第22条（本契約上の地位譲渡等の禁止）

甲及び乙は、相手方の書面による事前の承諾がない限り、本契約上の権利及び義務並びに本契約上の地位を第三者へ譲渡し、又は担保に供してはならない。

第23条（存続条項）

本契約が終了した場合でも、第10条から第15条、第17条、第18条、第20条、第21条3項から4項、及び、本条から第26条は有効に存続する。ただし、第14条については、本契約の終了後【2】年間に限り、その効力を有する。

第24条（準拠法）

本契約は、日本法を準拠法とする。

第25条（合意管轄）

本契約に関する一切の紛争については、法令に専属管轄の定めがある場合を除き、被告の住所地を管轄する地方裁判所又は簡易裁判所を第一審の専属的合意管轄裁判所とする。

第26条（協議）

甲及び乙は、本契約に定めがない場合及び本契約の条項の解釈について疑義が生じた場合は、民法その他の法令及び慣行に従い、誠意をもって協議し、解決するものとする。

本契約の成立を証するため本書2通を作成し、各自記名押印のうえ、各1通を保有する。

〇年〇月〇日

　　　　　　　　委託者（甲）
　　　　　　　　受託者（乙）

\end{contract}

\twocolumn

\tiny
\onecolumn
\begin{prompt}{Inference Prompt for \amendexp}{inference_prompt_amendexp}
\textbf{User prompt}\\\\
あなたは法律の専門家です。古い法令・新しい法令が与えられます。
まず本質的な変更点のみを列挙し（表記揺れ・語尾・軽微な用語統一・条ずれ等は無視）、続いて各変更が契約書へ与える影響を、根拠となる法令の条番号付きで列挙してください。
\\
【入力】\\
古い法令:
\{old\_statute\}
\\
新しい法令:
\{new\_statute\}
\\
\\
【作業規則】\\
- 実務影響が見込まれる**本質的変更**のみ抽出\\
- 表記の揺れ、句読点、順序入替、同義語置換など**軽微な変化は全て無視**\\
- 政府の参考資料が存在する場合は**そこに明記の変更点を優先**し、該当しない細部は出力しない\\
- 契約書への影響には、具体的な修正対象（条・項・別紙の種別など）と**根拠条番号（新旧）**を併記\\
- 変更点や影響がない場合は「（該当なし）」と明示\\
\\
【出力仕様】\\
- 出力は次のタグ内**のみ**に記載\\
- タグ内は**以下のコードブロック形式**を厳密に守る\\
\\
<OUTPUT>\\
\# 法令の変更点\\
- 変更点1\\
- 変更点2\\
\\
\# 契約書への影響\\
- 変更点1\\
- 変更点2\\
</OUTPUT>\\
\end{prompt}

\begin{prompt}{Inference Prompt for \statrev}{inference_prompt_statrev}
\textbf{User prompt}\\\\
あなたは経験豊富な法務専門家です。古い法令に基づく契約書を新しい法令に適合させてください。\\
【入力】\\
古い法令:\\
\{old\_statute\}\\
\\
新しい法令:\\
\{new\_statute\}\\
\\
修正対象の契約書:\\
\{contract\}\\
\\
【作業規則】\\
- 契約書の構造と形式は可能な限り維持\\
- 変更が必要な条文のみ修正（無関係な条文は出力しない）\\
- 条番号・見出しは元の表記を踏襲\\
\\
【出力仕様】\\
- 冒頭は **「修正後の契約書:」** の文字列から開始\\
- その直後に改行し、**差分がある条文のみ**を本文として出力\\
- 出力は次のタグで囲む。タグ外には一切出力しない\\
<OUTPUT>\\
修正後の契約書:\\
（ここに修正後の条文のみ）\\
</OUTPUT>\\
\end{prompt}

\begin{prompt}{Inference Prompt for \reqrev}{inference_prompt_reqrev}
\textbf{System prompt}\\\\
あなたは最高水準の法務専門家です。\\
契約書の修正において以下の品質基準を満たしてください：\\
\\
【必須要件】\\
- 指示された全ての修正を漏れなく実行\\
- 条項番号の整合性を完璧に維持\\
- 引用条項番号の適切な更新\\
- 法的に正確で明確な表現の使用\\
\\
【品質基準】\\
- 契約書の構造と論理性を保持\\
- 専門的で統一された文体\\
- 曖昧さのない明確な条文\\
\\
最高品質の修正結果を提供してください。\\
\\\textbf{User prompt}\\\\
以下の契約書に対して、指定された修正を行ってください。\\
\\
契約書:\\
\{contract\_text\}\\
\\
修正指示:\\
\{instruction\}\\
\\
修正された契約書全体を出力してください。\\
\end{prompt}

\begin{prompt}{Inference Prompt for \riskrev}{inference_prompt_riskrev}
\textbf{System prompt}\\\\
あなたは最高水準の法務専門家です。\\
契約書の修正において以下の品質基準を満たしてください：\\
\\
【必須要件】\\
- 指定された契約リスクを解消するための修正を漏れなく実行\\
- 条項番号の整合性を完璧に維持\\
- 引用条項番号の適切な更新\\
- 法的に正確で明確な表現の使用\\
\\
【品質基準】\\
- 契約書の構造と論理性を保持\\
- 専門的で統一された文体\\
- 曖昧さのない明確な条文\\
\\
最高品質の修正結果を提供してください。\\
\\\textbf{User prompt}\\\\
以下の契約書に対して、指定されたリスク対策の観点から修正を行ってください。\\
\\
契約書:\\
\{contract\_text\}\\
\\
修正指示:\\
\{instruction\}\\
\\
修正された契約書全体を出力してください。\\
\end{prompt}

\twocolumn

\tiny
\onecolumn

\begin{prompt}{Evaluation Prompt for \amendexp}{evaluation_prompt_amendexp}
\textbf{User prompt}\\\\
あなたは法律専門家による法令改正説明の評価を行う専門家です。以下の内容について評価してください。\\
\\
【評価対象】\\
元のプロンプト: \{original\_prompt\}\\
\\
LLMの回答: \{llm\_response\}\\
\\
【正解データ】\\
\{ground\_truth\}
\\
【評価基準】\\
下記の評価尺度に基づいて以下の観点から評価し、具体的な理由を述べてください：\\
\\
1. **改正の概要 - 主要な変更点の網羅性** (0-2点)  \\
   - `0`: 主要な変更点が網羅されておらず、重要な変更が多数漏れている\\  
   - `1`: 主要な変更点の一部は網羅されているが、不足がある  \\
   - `2`: 主要な変更点を十分に網羅しており、全体像を把握できる \\ 
\\
2. **改正の概要 - 変更点の説明精度** (0-2点)  \\
   - `0`: 変更点の説明に重大な誤りや不正確さがある\\  
   - `1`: 概ね正確だが、一部に不十分な説明や曖昧な表現がある\\  
   - `2`: 変更点の説明が正確で、法令解釈や表現も適切である  \\
\\
3. **改正の概要 - 不要内容の排除** (0-2点) \\ 
   - `0`: 不要な内容や本質的でない変更が多く含まれている  \\
   - `1`: 一部に不要な内容が含まれるが、全体として許容範囲  \\
   - `2`: 不要な内容は含まれず、簡潔で要点に絞られている  \\
\\
4. **契約書への影響 - 影響事項の網羅性** (0-2点)\\  
   - `0`: 契約書に影響する主要な事項を把握できていない \\ 
   - `1`: 一部の影響事項を把握しているが、不足がある  \\
   - `2`: 影響事項を十分に網羅し、実務的な影響も特定できている\\  
\\
5. **契約書への影響 - 影響説明の正確性** (0-2点)\\  
   - `0`: 影響説明に重大な誤りや不正確さがある  \\
   - `1`: 概ね正確だが、一部に不足や不明確な点がある\\  
   - `2`: 影響について正確に説明され、根拠や実務対応も明確である\\  
\\
6. **契約書への影響 - 不要内容の排除** (0-2点)\\  
   - `0`: 契約書に関係のない不要な内容が多く含まれている\\  
   - `1`: 一部に不要な内容が含まれるが、全体として許容範囲\\  
   - `2`: 不要な内容を含まず、実務に直結した要点に絞られている\\  
\end{prompt}

\begin{prompt}{Evaluation Prompt for \statrev}{evaluation_prompt_statrev}
\textbf{User prompt}\\\\
あなたは法改正による契約書修正の評価を行う専門家です。以下の内容について評価してください。\\
\\
【評価対象】\\
元のプロンプト: \{original\_prompt\}
\\
LLMの回答: \{llm\_response\}
\\
【正解データ】\\
\{ground\_truth\}\\
\\
【評価基準】\\
下記の評価尺度に基づいて以下の観点から評価し、具体的な理由を述べてください：\\
\\
1. **修正内容の新法令準拠性** (0-2点)\\  
   - `0`: 新法令に全く準拠していない／誤解を招く修正をしている  \\
   - `1`: 新法令の一部は反映されているが、不十分または誤りがある  \\
   - `2`: 新法令に完全に準拠し、解釈や反映も適切である  \\
\\
2. **契約書構造（条番号など）が整合している** (0-1点)\\
   - `0`: 契約書構造の整合性が失われている\\
   - `1`: 契約書構造は整合している\\
\\
3. **不要な修正がされていないか** (0-1点)\\
   - `0`: 不要な修正がされている\\
   - `1`: 不要な修正がされていない\\
\\
4. **専門用語の正確性** (0-2点)\\
   - `0`: 実務上重大な誤りのある専門用語の使用をしている  \\
   - `1`: 実務上問題はないが、不自然または誤解を招く専門用語がある\\  
   - `2`: 専門用語が正確かつ自然に使用されている  \\
\\
5. **契約書言い回しの適切性** (0-2点)\\
   - `0`: 契約書的な言い回しが全く適切でなく、口語的・不明確である\\  
   - `1`: 契約書的な言い回しが一部適切でない、または曖昧な表現がある\\  
   - `2`: 契約書的な言い回しが適切で、明確かつ簡潔である  \\
\end{prompt}

\begin{prompt}{Evaluation Prompt for \reqrev}{evaluation_prompt_reqrev}
\textbf{User prompt}\\\\
あなたは契約書修正の評価を行う法務専門家です。以下の内容について評価してください。\\
\\
【評価対象】\\
元の契約書: \{original\_prompt\}\\
\\
修正指示: \{instruction\}\\
\\
LLMによる修正後契約書: \{llm\_response\}\\
\\
【正解データ】\\
\{ground\_truth\}\\
\\
【評価基準】\\
下記の評価尺度に基づいて以下の観点から評価し、具体的な理由を述べてください：\\
\\
1. **指示通りの修正がされているか** (0-2点)\\
   - `0`: 要件が全く満たされていない\\
   - `1`: 要件が一部満たされているが、一部満たされていない\\
   - `2`: 要件が全て満たされている\\
\\
2. **契約書構造（条番号など）が整合している** (0-1点)\\
   - `0`: 契約書構造の整合性が失われている\\
   - `1`: 契約書構造は整合している\\
\\
3. **指示されていない不要な修正がされていないか** (0-1点)\\
   - `0`: 指示されていない不要な修正がされている\\
   - `1`: 指示されていない不要な修正がされていない。条番号や項番号、引用条項に関する書式の変更は許容される。修正指示を拡大解釈して修正している場合も許容される。\\
\\
4. **専門用語が正しく使用されているか** (0-2点)\\
   - `0`: 実務上問題になる誤った専門用語の使い方をしている\\
   - `1`: 実務上問題はないが、専門用語が適切に使用されていない\\
   - `2`: 専門用語に関して違和感なく自然な表現が用いられている。曖昧な表現や広範な表現については許容される。\\
\\
5. **契約書的な言い回しが適切に使用されているか** (0-2点)\\
   - `0`: 契約書的な言い回しが全く適切でない\\
   - `1`: 契約書的な言い回しが一部適切でない\\
   - `2`: 契約書的な言い回しが全て適切に用いられている\\
\end{prompt}

\begin{prompt}{Evaluation Prompt for \riskrev}{evaluation_prompt_riskrev}
\textbf{User prompt}\\\\
あなたは契約書修正の評価を行う法務専門家です。以下の内容について評価してください。\\
\\
【評価対象】\\
元の契約書: \{original\_prompt\}\\
\\
修正指示: \{instruction\}\\
\\
LLMによる修正後契約書: \{llm\_response\}\\
\\
【正解データ】\\
\{ground\_truth\}\\
\\
【評価基準】\\
下記の評価尺度に基づいて以下の観点から評価し、具体的な理由を述べてください：\\
\\
1. **指示通りの修正がされているか** (0-2点)\\
   - `0`: リスクが全く解消されていない\\
   - `1`: リスクが一部解消されているが、一部解消されていない\\
   - `2`: リスクが全て解消されている\\
\\
2. **契約書構造（条番号など）が整合している** (0-1点)\\
   - `0`: 契約書構造の整合性が失われている\\
   - `1`: 契約書構造は整合している\\
\\
3. **不要な修正がされていないか** (0-1点)\\
   - `0`: 不要な修正がされている\\
   - `1`: 不要な修正がされていない\\
\\
4. **専門用語が正しく使用されているか** (0-2点)\\
   - `0`: 実務上問題になる誤った専門用語の使い方をしている\\
   - `1`: 実務上問題はないが、専門用語が適切に使用されていない\\
   - `2`: 専門用語に関して違和感なく自然な表現が用いられている\\
\\
5. **契約書的な言い回しが適切に使用されているか** (0-2点)\\
   - `0`: 契約書的な言い回しが全く適切でない\\
   - `1`: 契約書的な言い回しが一部適切でない\\
   - `2`: 契約書的な言い回しが全て適切に用いられている\\
\end{prompt}

\twocolumn

\end{document}